\documentclass{article}

\PassOptionsToPackage{numbers, compress}{natbib}

\usepackage[main, final]{neurips_2026}
\makeatletter
\renewcommand{\@notice}{}
\makeatother

\usepackage[utf8]{inputenc} 
\usepackage[T1]{fontenc}    
\usepackage{hyperref}       
\usepackage{url}            
\usepackage{booktabs}       
\usepackage{amsfonts}       
\usepackage{nicefrac}       
\usepackage{microtype}      
\usepackage{xcolor}         
\usepackage{graphicx}
\usepackage{multirow}
\usepackage{wrapfig}
\usepackage{orcidlink}
\usepackage{tabularx}
\usepackage{array}
\usepackage{hyperref}
\usepackage{amsmath}
\usepackage{cleveref}
\usepackage{subcaption}
\usepackage{natbib}
\usepackage{algorithm}
\usepackage{algpseudocode}

\title{SubdivAR: Autoregressive Next-Scale Prediction for Neural Mesh Subdivision}

\author{
\textbf{Huipeng Guo}$^{1}$ \quad
\textbf{Zikai Song}$^{1}$ \quad
\textbf{Hang Long}$^{1}$ \quad
\textbf{Jielei Zhang}$^{1}$ \quad
\textbf{Wenbing Li}$^{1}$ \quad
\textbf{Junkai Lin}$^{1}$ \\
\textbf{Tianhao Zhao}$^{1}$ \quad
\textbf{Jinshen Zhang}$^{1}$ \quad
\textbf{Tianle Guo}$^{1}$ \quad
\textbf{Wei Yang}$^{1\dag}$  \\
Huazhong University of Science and Technology \\
}

\begin{document}

\maketitle
\begin{figure}[htbp]
    \centering
    \includegraphics[width=1.0\linewidth]{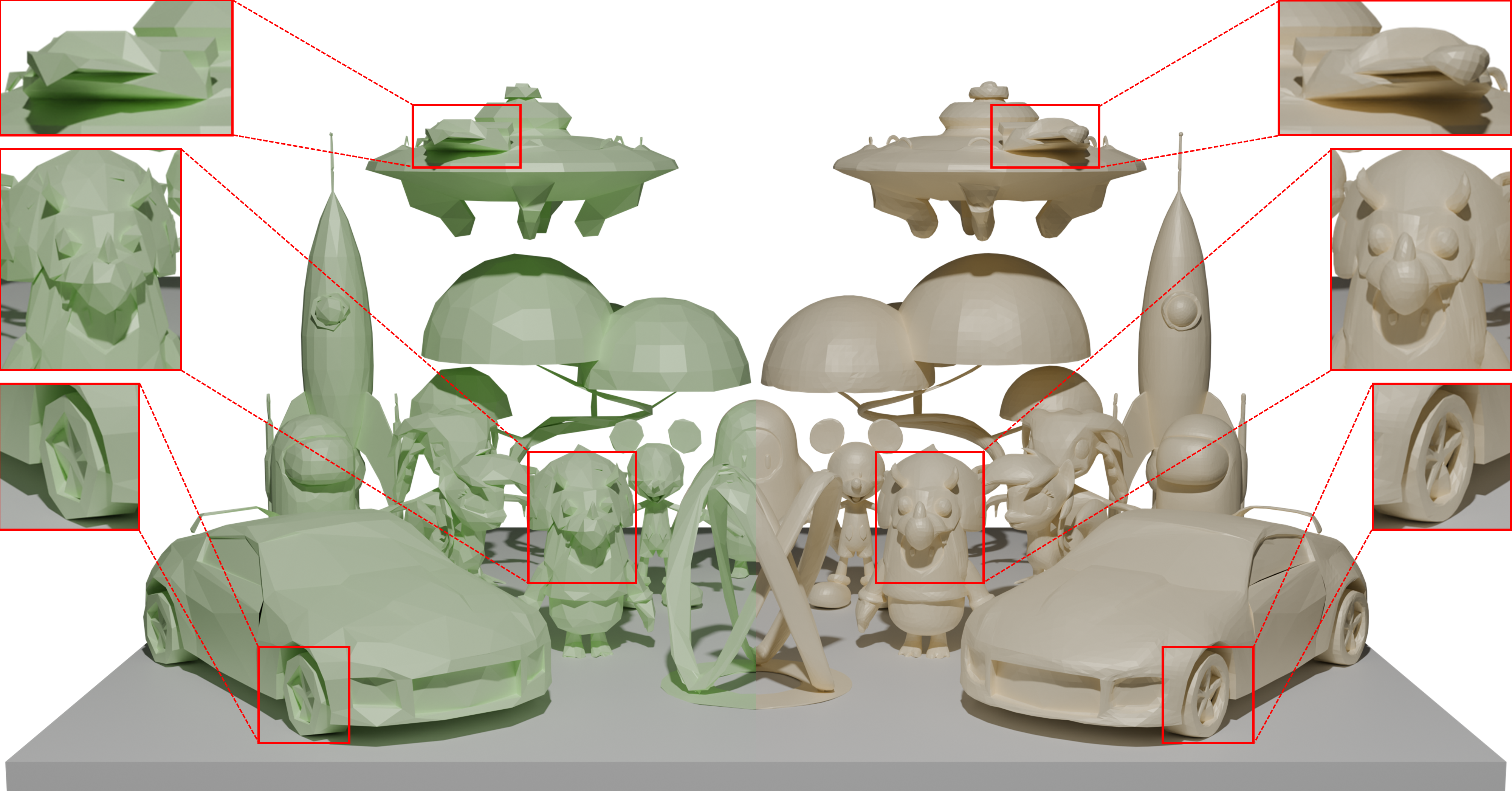} 
    \caption{\textbf{High-fidelity mesh subdivision achieved by \textbf{SubdivAR}}. Given a coarse input mesh (left, green), our model produce refined surfaces with faithful semantic and geometric details.}
    \label{fig:teazer}
\end{figure}

\begin{abstract}
\vspace{-1em}
Mesh subdivision is a fundamental operation for converting coarse, editable meshes into high-resolution surfaces, with broad applications in digital asset creation. Classical rule-based schemes rely on fixed local refinement rules and often produce over-smoothed surfaces. Recent neural subdivision methods improve detail synthesis, but remain constrained by local modeling and exhibit limited generalizability.
We present \textbf{SubdivAR}, a neural mesh subdivision framework based on our proposed \textbf{Mesh Autoregressive Representation} (MAR). MAR arranges meshes at different subdivision levels into an ordered scale sequence, reformulating subdivision as autoregressive next-scale prediction. To support this formulation, we introduce a Hybrid Topology-Aware Transformer that combines global semantic attention with topology-constrained local feature aggregation. SubdivAR adopts a next-scale coordinate prediction paradigm, regressing vertex offsets at each refinement stage to preserve subdivision topology while recovering fine-grained geometric details.
To enable reliable learning, we construct \textbf{FII-40K}, a curated dataset of nearly 40,000 high-quality meshes with multi-level subdivision supervision. Experiments show that SubdivAR outperforms state-of-the-art baselines, reducing Hausdorff Distance and Chamfer Distance by 18.8\% and 14.2\%, respectively, and demonstrates strong robustness on complex open-surface geometries.
\end{abstract}

\section{Introduction}
\label{sec:intro}
\vspace{-1em}
In professional animation, game development, and digital asset creation, mesh subdivision provides a standard coarse-to-fine modeling paradigm: artists first manipulate a sparse control cage as a lightweight proxy for modeling, rigging, and editing, and then refine it into a high-resolution surface. Unlike unstructured 3D representations, such as point clouds or voxels, which are widely used in 3D super-resolution tasks~\cite{yu2018pu, convocc} or as intermediate priors for decoupled mesh reconstruction~\cite{siddiqui2024meshgpt, zhao2026lato}, subdivision explicitly preserves mesh connectivity and topological consistency throughout refinement. This structural regularity is essential for retaining the underlying modeling intent required by downstream production pipelines. However, classical rule-based subdivision schemes~\cite{catmull, loop, kobbelt2000sqrt3, zorin1996interpolating} rely on fixed, geometry-agnostic linear stencils. Although robust and easy to implement, these stationary rules provide limited adaptability to complex geometry and often over-smooth surfaces, erasing sharp features and failing to recover high-frequency details.

To overcome the rigidity of hand-crafted subdivision rules, recent neural subdivision methods attempt to replace fixed stencils with learnable refinement operators. Representative approaches, including Neural Subdivision (NeuralSubdiv)~\cite{NeurSu}, Neural Mesh Refinement (NMR)~\cite{NeurMeRe}, and Graph Neural Subdivision (GNS)~\cite{NeurSuGNN}, learn vertex displacements from local neighborhoods or graph neural networks. Despite encouraging progress, existing methods remain limited by major bottlenecks. First, they face a structural modeling gap: most architectures are dominated by local message passing and therefore struggle to capture the global semantic context needed for coherent, shape-aware refinement. Increasing the depth of purely local graph networks does not fully resolve this issue, as repeated-message passing can lead to feature over-smoothing~\cite{li2018deeperinsightsgraphconvolutional}, echoing the geometric over-smoothing observed in classical subdivision rules. Second, prior methods primarily assume watertight or closed topologies, limiting their applicability to the open-surface meshes commonly found in modern asset repositories.
At last, they suffer from a \textbf{data quality gap}: training pairs generated by uncurated synthetic decimation often contain irregular connectivity, geometric artifacts, and unreliable coarse-to-fine supervision, which weakens the learned refinement prior. 

We address these challenges with \textbf{SubdivAR}, a neural mesh subdivision framework that combines a novel mesh autoregressive multi-scale formulation. Inspired by the success of Visual Autoregressive (VAR) modeling~\cite{var} on image pyramids, we reformulate neural subdivision as a \textbf{Next-Scale Coordinate Prediction} problem: meshes at different subdivision levels are organized into an ordered scale sequence, and the model progressively predicts vertex offsets from a coarser level to the next finer level. This formulation enables SubdivAR to preserve the prescribed subdivision topology while recovering fine-grained geometric details in a hierarchical manner.
To overcome the structural limitations of local refinement, we introduce a \textbf{Hybrid Topology-Aware Transformer} that integrates global semantic attention with topology-constrained local feature aggregation. Importantly, SubdivAR naturally supports both closed and open surfaces, making it better aligned with real-world asset collections. 
We first develop a data curation pipeline that filters subdivision pairs according to geometric fidelity and topological integrity, resulting in \textbf{FII-40K}, a curated benchmark containing nearly 40,000 high-quality meshes.
Through extensive experiments, SubdivAR significantly outperforms state-of-the-art baselines, and demonstrates promising generalization to varied coarse inputs, as shown in \cref{fig:on_generative}, suggesting its potential compatibility with neural simplification~\cite{nms}, generative mesh workflows~\cite{siddiqui2024meshgpt}, and topology-preserving decimation techniques~\cite{xu2024cwf} that produce regularized coarse topologies.

In summary, our contributions are threefold. 
\begin{itemize}
    \vspace{-0.5em}
    \item We introduce a \textbf{Mesh Autoregressive Representation} paradigm that formulates subdivision as autoregressive hierarchical refinement across resolution levels. 
    \item We propose a \textbf{Hybrid Topology-Aware Transformer} that jointly captures global shape semantics and local topological constraints for coherent mesh refinement, successfully tackle non-watertight surfaces.
    \item We construct \textbf{FII-40K}, a large-scale curated benchmark designed to provide clean and reliable supervision for neural mesh subdivision.
\end{itemize}

\section{Related Work}
\label{sec:relwork}
\paragraph{Mesh Subdivision and Refinement.}
Traditional subdivision schemes generate smooth surfaces by recursively defining limit surfaces from coarse polygonal meshes~\cite{loop, catmull, doo1978behavior, kobbelt2000sqrt3, zorin1996interpolating, hoppe1994piecewise}. However, these heuristic rules are largely geometry-agnostic, which often leads to over-smoothed sharp features and limited recovery of non-trivial local details. To overcome the rigidity of fixed stencils, learning-based approaches such as Neural Subdivision (NS)~\cite{NeurSu}, Neural Mesh Refinement (NMR)~\cite{NeurMeRe}, and Graph Neural Subdivision (GNS)~\cite{NeurSuGNN} replace hand-crafted rules with trainable operators. Despite their progress, these methods are typically driven by local neighborhood structures and therefore may overlook global shape context and high-level geometric intent. In particular, NS~\cite{NeurSu} and NMR~\cite{NeurMeRe} are designed for closed manifold meshes and cannot directly handle open boundaries. Recent works have further explored progressive transmission~\cite{neuralprogress} and implicit subdivision representations~\cite{INDONS}, yet achieving high-fidelity refinement while preserving strict topological intent remains challenging.

\paragraph{3D Super-Resolution.}
Super-resolution in 3D has also been studied through point cloud upsampling~\cite{yu2018pu, qian2021pu}, voxel or implicit upsampling~\cite{convocc, shen2021deep, park2019deepsdf}, and generative mesh reconstruction~\cite{siddiqui2024meshgpt, lin2025meshripple, zhao2026lato}. These methods can increase geometric density or synthesize detailed shapes, but they often do not preserve the explicit edge connectivity required by professional modeling workflows. A possible pipeline is to first generate dense unstructured geometry, such as point clouds or high-resolution implicit fields, and then convert it into a regularized mesh using a generative mesh model. However, this decoupled process can deviate from the precise geometry and topology of the original base mesh, introducing artifacts such as non-manifold holes or double-layer surfaces. Moreover, autoregressive mesh generation frameworks such as ARMesh~\cite{ARMesh} can suffer from long inference latency and limited output resolution due to sequential sampling. In contrast, our method treats subdivision as structure-preserving refinement, aiming to recover high-resolution geometry while maintaining the topological intent of the input mesh.

\paragraph{Geometric Feature Extraction.}
Extracting expressive features from irregular 3D structures is central to mesh and point-based learning. Early specialized mesh convolutions~\cite{meshcnn, subdivnet}, point-based context aggregators~\cite{qi2017pointnet, qi2017pointnet++}, and graph neural networks~\cite{monti2017monet, verma2018feastnet, pfaff2021meshgraphnet, gong2019spiralnet} provide effective mechanisms for local geometric reasoning, but purely local message passing can suffer from over-smoothing as network depth increases~\cite{li2018deeperinsightsgraphconvolutional}. Recent attention-based 3D backbones~\cite{Zhao_2021_ICCV, clay, trellis, instantmesh2024, zhao2025hunyuan3d} improve global context modeling, while 2D image super-resolution has similarly moved toward multi-scale Transformer designs~\cite{Liang_2021_ICCV, chen2023activating}. In particular, Visual Autoregressive (VAR) modeling~\cite{var} reformulates generation as next-scale prediction over image pyramids~\cite{vaswani2017attention}, which is naturally aligned with progressive mesh subdivision. Inspired by this hierarchy, our approach combines global semantic attention with local manifold-constrained features and treats subdivision as a multi-scale coordinate regression problem, preserving structural consistency from the base mesh to fine geometric details.

\section{Method}
\label{sec:method}
In this section, we present the proposed \textbf{SubdivAR} framework. We first introduce \textbf{FII-40K}, a curated dataset consisting varying level of mesh subdivisions, constructed through systematic data collection, filtering, and cleaning. We then describe the SubdivAR architecture, which combines topology-aware hybrid attention with auxiliary boundary features for vertex feature extraction, and performs hierarchical vertex-offset regression under a next-scale prediction formulation.

\subsection{The FII-40K Dataset} 

\begin{wrapfigure}[10]{r}{0.4\textwidth} 
    \vspace{-3em} 
    \centering
    \includegraphics[width=\linewidth]{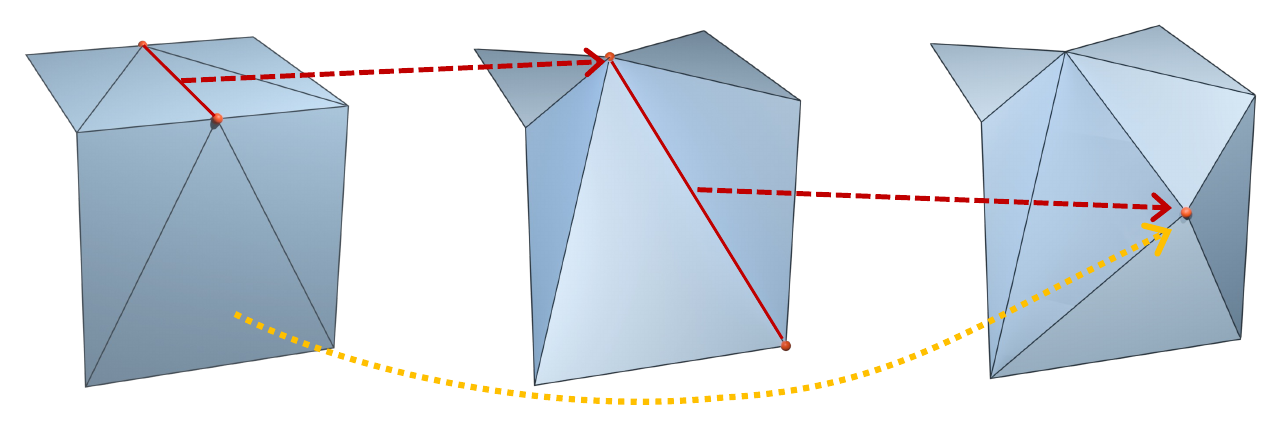} 
    \vspace{-0.5em} 
    \caption{\textbf{Sharp Feature Collapse.} The shape distortion is attributed to the failure in preserving sharp features during reparameterization.}
    \label{fig:sharp_feature_collapse}
    \vspace{-1em} 
\end{wrapfigure}

Neural mesh subdivision is limited by the scarcity of paired coarse-fine mesh data. To address this issue, we construct the \textbf{FII-40K} dataset, which contains approximately 40,000 meshes with multi-level subdivision supervision. We begin by following the pipeline of Neural Subdivision~\citep{NeurSu}: each high-resolution mesh is simplified to a target face count through edge collapse and then aligned with the original surface via conformal mapping. The positions of newly generated vertices are determined from the conformal mapping of neighboring vertices. Although this process provides large amount of candidates, it does not account for several important failure cases. For meshes with sharp features or complex local structures, edge collapse may cause severe shape distortion, irregular connectivity, and unreliable coarse-to-fine correspondences. Such artifacts introduce noisy supervision and hinder scalable learning. We therefore apply a systematic data curation pipeline to retain pairs with reliable geometry and topology.


\begin{table}[htbp]
\centering
\caption{Detailed metrics for data curation. The $T_i$ values are optimized via \cref{eq:threshold_opt} to balance recall and precision. Only key metrics are listed here.}
\label{tab:metrics_details}
\begin{tabular}{@{}lccccc@{}}
\toprule
Category & Metric ($m_i$) & Scope & Objective & $T_i$ & Valid Condition \\ \midrule
Fidelity & HD & $M_{sub} \leftrightarrow M_{orig}$ & Accuracy & 0.03 & $m_i < T_i$ \\
                  & CD & $M_{sub} \leftrightarrow M_{orig}$ & Accuracy & 0.0045 & $m_i < T_i$ \\ \midrule
Integrity & Narrow Face Ratio & $M_{sub}$ & Topology & 0.1 & $m_i < T_i$ \\
                  & Self-intersection & $M_{sub} \leftarrow M_{orig}$ & Topology & 0.08 & $m_i < T_i$ \\ \midrule
Informativeness & Curvature Ratio & $M_{orig}$ & Diversity & 0.65 & $m_i > T_i$ \\ \bottomrule
\end{tabular}
\end{table}

\begin{figure}[t]
    \centering
    \includegraphics[width=1\linewidth]{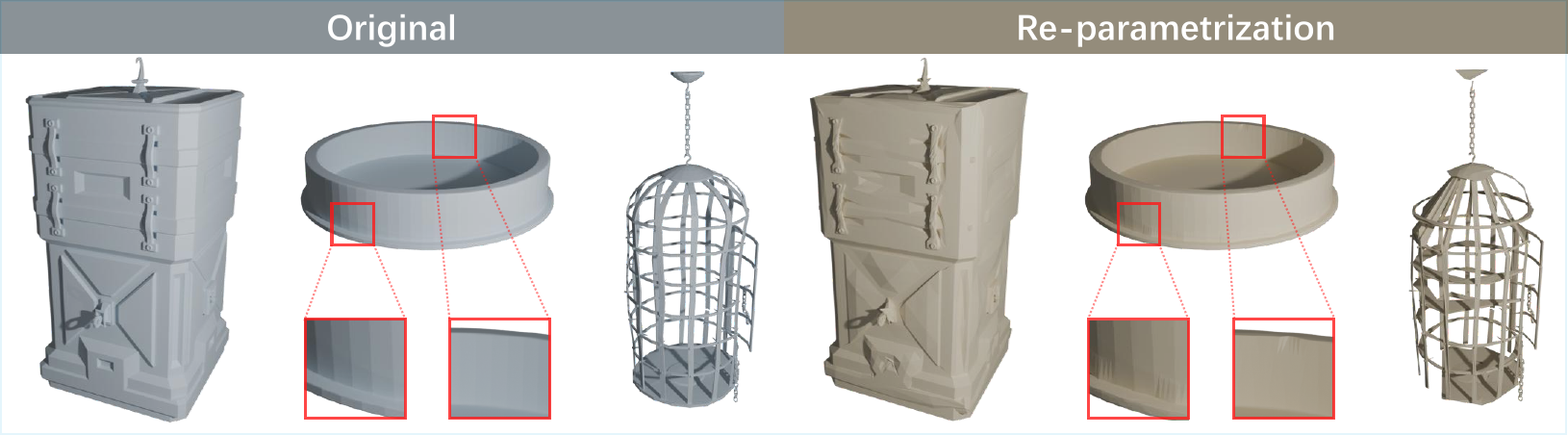}
    \caption{\textbf{Failure Modes.} Comparing the raw mesh (left) with the reparameterized result (right), we identify two primary failure modes: significant shape distortion with a loss of sharp features(Figure~\ref{fig:sharp_feature_collapse}), and degraded mesh quality characterized by self-intersections and degenerated sliver faces.}
    \label{fig:failure_modes}
\end{figure}

\paragraph{Filtering Metric} 
Our analysis reveals two primary failure modes during data construction, as illustrated in Figure~\ref{fig:failure_modes}, as well as an additional criterion related to training efficiency. To systematically curate the generated data, we group the selection metrics $m_i$ into three dimensions, as summarized in Table~\ref{tab:metrics_details}. First, \textbf{Fidelity} measures geometric distortion introduced during simplification, especially when sharp features are collapsed, and is evaluated using Hausdorff Distance (HD) and Chamfer Distance (CD). Second, \textbf{Integrity} assesses the quality of mesh connectivity, targeting irregular topological layouts, or ``messy wiring'', that often result from failed UV mapping. This dimension is quantified by the Narrow Face Ratio and Self-intersection Rate. Third, \textbf{Informativeness} improves training efficiency by filtering out trivial samples dominated by flat regions, using the Curvature Feature Ratio computed on the original mesh.


\paragraph{Thresholding.}
To distinguish qualified coarse-fine pairs from failed constructions, we manually curate a calibration set containing 100 successful and 100 failed samples. We then compute the empirical distribution of each selected metric for both groups, as shown in Figure~\ref{fig:metric}. Since each metric corresponds to a specific failure mode or quality dimension, we adopt an independent thresholding strategy. For each metric $m_i$, the threshold $T_i$ is selected by
\begin{equation} 
T_i = \arg \max_{\tau} \left\{ \omega \cdot \mathcal{R}_{\mathrm{succ}}(\tau) + (1 - \omega) \cdot \mathcal{E}_{\mathrm{fail}}(\tau) \right\}, 
\label{eq:threshold_opt} 
\end{equation} 
where $\mathcal{R}_{\mathrm{succ}}(\tau)$ denotes the recall of successful samples under threshold $\tau$, and $\mathcal{E}_{\mathrm{fail}}(\tau)$ denotes the exclusion rate of failed samples. We set a relatively large weight $\omega$ to prioritize data coverage while still rejecting low-quality samples.
After raw meshes are processed by the standardized preprocessing pipeline described in \cref{sec:datapreocess}, we apply the learned thresholds for automatic filtering. A sample $x$ is included in the final curated training set $\mathcal{D}_{\mathrm{final}}$ if and only if it satisfies all independent constraints across the three quality dimensions:
\begin{equation} 
x \in \mathcal{D}_{\mathrm{final}} \iff 
\forall i \in \{1, \dots, n\},\ 
\mathrm{valid}(m_i(x), T_i), 
\label{eq:final_filter} 
\end{equation} 
where $\mathrm{valid}(\cdot)$ denotes the metric-specific validity check, such as $<$ or $>$, according to the criterion defined in \cref{tab:metrics_details}, with thresholds $T_i$ optimized by \cref{eq:threshold_opt}.

After careful filtering and validation, we construct \textbf{FII-40K}, a curated dataset of high-quality coarse-fine mesh pairs that provides reliable supervision for training neural mesh subdivision models.


\subsection{Mesh Subdivision through Next Scale Prediction}
\begin{figure}[t]
    \centering
    \begin{subfigure}[b]{1\textwidth}
        \centering
        \includegraphics[width=\linewidth]{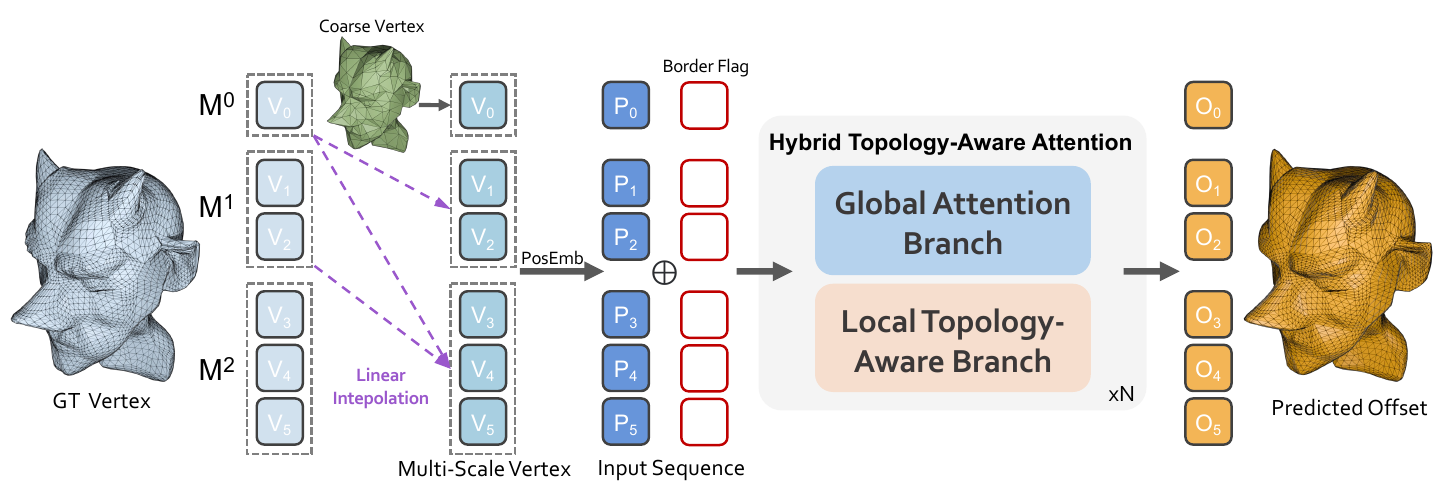}
        \caption{\textbf{Next scale coordinate prediction:} The input sequence is constructed from the coarse mesh and the linear subdivision of the previous ground-truth (GT). Specifically, we adopt the Loop subdivision scheme to determine the topological connectivity of the new midpoints, while their vertex positions are calculated via linear interpolation. These inputs are processed by concatenating its positional encoding with boundary features, followed by a hybrid attention mechanism to predict vertex offsets.}
        \label{fig:nextscale} 
    \end{subfigure}

    \begin{subfigure}[b]{1\textwidth}
        \centering
        \includegraphics[width=\linewidth]{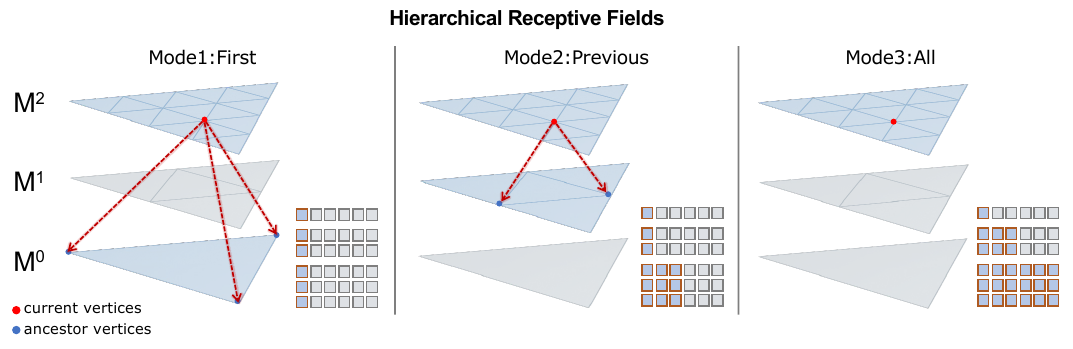}
        \caption{\textbf{Detailed view of the receptive-field-constrained hybrid attention module}.
When local attention is employed, each vertex locates its ancestor based on the chosen mode (first, previous, or all) and searches for neighbors within the corresponding hierarchy.
In the global attention mode, the receptive field is constrained by a block-diagonal mask, as illustrated in the figure. } 
        \label{fig:Receptivemode} 
    \end{subfigure}
    
    \caption{The pipeline of our SubdivAR framework.}
    \label{fig:arch}
\end{figure}

Given a datapoint from \textbf{FII-40K}, we construct a unified sequence $\mathbf{X}$ by concatenating vertices across all subdivision levels. For a target subdivision depth $L$, the sequence is defined as
\begin{equation}
    \mathbf{X} = \left[ \mathcal{V}^{(0)}, \mathcal{V}^{(1)}_{new}, \dots, \mathcal{V}^{(L)}_{new} \right],
\end{equation}
where $\mathcal{V}^{(0)}$ denotes the vertices of the coarsest mesh. For each level $l \geq 1$, the newly introduced vertices $\mathcal{V}_{\mathrm{new}}^{(l)}$ are initialized by linear interpolation at the edge midpoints of the ground-truth mesh at level $l-1$. Each vertex $\mathbf{p}_i$ in the sequence is represented by an augmented feature vector:
\begin{equation}
    \mathbf{f}_i = \gamma(\mathbf{p}_i) \oplus \mathbf{b}_i,
\end{equation}
where $\gamma(\cdot)$ denotes the sinusoidal positional encoding~\cite{nerf}, which provides high-frequency geometric embeddings, and $\oplus$ denotes feature concatenation. To handle boundary constraints in open surfaces, we further concatenate a binary boundary indicator $\mathbf{b}_i$, which provides manifold-aware context for robust refinement.


\paragraph{Hybrid Topology-Aware Attention}
Previous methods~\cite{NeurSu, NeurMeRe} rely on local pooling or graph convolutions, limiting their ability to capture global context for distinguishing both object-level semantics and fine-grained parts. We therefore introduce a hybrid attention mechanism that combines global semantic reasoning with topology-aware local feature aggregation.
Specifically, we employ a global cross-attention branch to capture long-range dependencies and global geometric priors, enabling each vertex to aggregate contextual features from the entire mesh. This provides shape-level semantic cues that are difficult to infer from local neighborhoods alone. However, unrestricted global attention may introduce spurious interactions between vertices that are spatially close but topologically disconnected, leading to geometric artifacts or hallucinated details. To mitigate this issue, we introduce a \textbf{Topology-Constrained Cross-Attention} branch, which follows the same attention formulation as the global branch but restricts the context tokens to the topological neighborhood $\mathcal{N}(i)$ of each query vertex. This design preserves local manifold consistency while retaining the benefits of global semantic context. The local feature $\mathbf{z}_{l,i}$ is 
%
%
\begin{equation}
    \mathbf{z}_{l,i} = \sum_{j \in \mathcal{N}(i)} \left( \frac{\exp\left(\frac{(\mathbf{W}'_q \mathbf{x}_i)(\mathbf{W}'_k \mathbf{c}_j)^\top}{\sqrt{d_k}}\right)}{\sum_{m \in \mathcal{N}(i)} \exp\left(\frac{(\mathbf{W}'_q \mathbf{x}_i)(\mathbf{W}'_k \mathbf{c}_m)^\top}{\sqrt{d_k}}\right)} \right) (\mathbf{W}'_v \mathbf{c}_j)
\end{equation}
where $\mathbf{W}'_q, \mathbf{W}'_k, \mathbf{W}'_v$ are separate learnable projections for local feature extraction. By performing attention within the strictly defined manifold neighborhood $\mathcal{N}(i)$, the model remains sensitive to local curvatures and sharp geometric features, ensuring that the refinement process is guided by the underlying mesh topology rather than mere spatial proximity.
\paragraph{Next-Scale Coordinate Prediction}
\label{nscp}
In VAR modeling (\citet{var}), image generation is formulated as a multi-scale conditional probability distribution:
\begin{equation}
    p(X) = \prod_{k=1}^K p(r_k \mid r_1, r_2, \dots, r_{k-1})
\end{equation}
where $r_k$ represents the feature map at the $k$-th resolution scale. We observe that mesh subdivision naturally aligns with this multi-scale framework, where each scale corresponds to the set of new vertices $\mathcal{V}_{new}^{(s)}$ generated at the $s$-th subdivision level. Our hierarchical ``Next-Scalell approach enables efficient ``train once, predict multiplell refinement by regressing offsets only for new vertices. By fixing predecessors, it drastically reduces sequence length compared to full-mesh re-processing. Unlike flat cross-attention method \cite{3dshape}, this design captures critical inter-level dependencies, ensuring superior geometric fidelity with significantly lower computational overhead.

\paragraph{Hierarchical Receptive Fields.}
A critical factor in this architecture is the scope of the receptive field for new vertices. We define three distinct modes for the global attention context (see Figure~\ref{fig:Receptivemode}): \textbf{1) First}, where the query vertex only observes the coarsest base mesh $\mathcal{V}^{(0)}$; \textbf{2) Previous}, where the query observes vertices from $\mathcal{V}^{(0)}$ to $\mathcal{V}^{(s-1)}$; and \textbf{3) All}, where the query observes all vertices from level $0$ up to the current level $s$.
Correspondingly, the neighborhood $\mathcal{N}_{local}(v_i^{(s)})$ for the Local Topology-Aware Branch is based on a level-specific ancestor querying process. We first define a generalized projection function $\phi_{k}(v^{(s)})$, which maps a vertex $v^{(s)}$ at level $s$ to its set of ancestor vertices at a target level $k$ ($0 \le k \le s$). 
Let $v_{i}^{(s)}$ be a new vertex generated from the edge $(u^{(s-1)}, w^{(s-1)})$. Its ancestor set at level $k$ is recursively defined as:
\begin{equation}
    \phi_k(v_i^{(s)}) = 
    \begin{cases} 
    \{ v_i^{(s)} \}, & k = s \\
    \{ u^{(s-1)}, w^{(s-1)} \}, & k = s-1 \\
    \phi_{0}(u^{(s-1)}) \cup \phi_{0}(w^{(s-1)}), &  k = 0 \text{ and } s \neq 1
    \end{cases}
\end{equation}
where $\phi_{0}(\cdot)$ ultimately returns the constituent vertices in the base mesh $\mathcal{M}^{(0)}$. 
For a given vertex set $V$, we define its neighborhood at level $k$ as the union of individual neighborhoods, i.e., $\mathcal{N}^{(k)}(V) = \bigcup_{v \in V} \mathcal{N}^{(k)}(v)$. The local neighborhood mapping is then elegantly unified as finding the ancestors at level $k$ followed by extracting their neighborhoods at that same level:
\begin{equation}
    \mathcal{N}_{local}(v_i^{(s)}) = 
    \begin{cases} 
    \mathcal{N}^{(s)}\big( \phi_{s}(v_i^{(s)}) \big), & \text{mode: \textit{All} } (k=s) \\
    \mathcal{N}^{(s-1)}\big( \phi_{s-1}(v_i^{(s)}) \big), & \text{mode: \textit{Previous} } (k=s-1) \\
    \mathcal{N}^{(0)}\big( \phi_{0}(v_i^{(s)}) \big), & \text{mode: \textit{First} } (k=0)
    \end{cases}
\end{equation}
These modes balance global structure and local detail. Empirically, the optimal selection depends on data scale: \textit{First} mode regularizes small datasets to prevent overfitting, while \textit{All} mode excels on larger datasets by capturing intricate multi-level dependencies.

\paragraph{Training Objectives and Stability Strategies}
Subdivision training often converges to trivial zero-offset mappings due to the highly imbalanced distribution of vertex displacements. To counter this, we employ a staged supervision strategy. During the initial phase ($t \le T_{init}$), we deactivate the normal loss and exclude the base mesh vertices $\mathcal{V}^{(0)}$ from training, supervising only the newly generated vertices $\mathcal{V}_{new} = \mathcal{V}^{(L)} \setminus \mathcal{V}^{(0)}$. This forces the model to prioritize non-trivial geometric refinements over identity mapping. The total loss $\mathcal{L}$ is:
\begin{equation}
    \mathcal{L} = \mathcal{L}_{coord} + \mathbb{1}_{t > T_{init}} \cdot \lambda_{norm} \mathcal{L}_{norm}
\end{equation}
where $\mathcal{L}_{coord} = \frac{1}{|\mathcal{V}_{train}|} \sum_{i \in \mathcal{V}_{train}} \| \Delta \mathbf{p}_i - \lambda (\mathbf{p}_i^{gt} - \mathbf{p}_i^{in}) \|^2$. $\mathcal{V}_{train}$ switches from $\mathcal{V}_{new}$ to $\mathcal{V}^{(L)}$ after $T_{init}$. We set $\lambda = 100$ to amplify infinitesimal displacements (typically $< 0.03$), preventing gradient vanishing and enhancing sensitivity to subtle details \cite{triposg}. In the later stage, we activate the face normal loss $\mathcal{L}_{norm}$ to ensure global surface consistency:
\begin{equation}
    \mathcal{L}_{norm} = \frac{1}{|\mathcal{F}|} \sum_{f \in \mathcal{F}} \left\| \frac{(\mathbf{v}_{f,1} - \mathbf{v}_{f,0}) \times (\mathbf{v}_{f,2} - \mathbf{v}_{f,0})}{\| (\mathbf{v}_{f,1} - \mathbf{v}_{f,0}) \times (\mathbf{v}_{f,2} - \mathbf{v}_{f,0}) \|} - \mathbf{n}_f^{gt} \right\|^2
\end{equation}
This joint refinement across all vertices ensures high-fidelity surface reconstruction once the model has captured meaningful geometric offsets.
\section{Experiments}
\begin{figure}[htbp]
    \centering
    \includegraphics[width=1.0\linewidth]{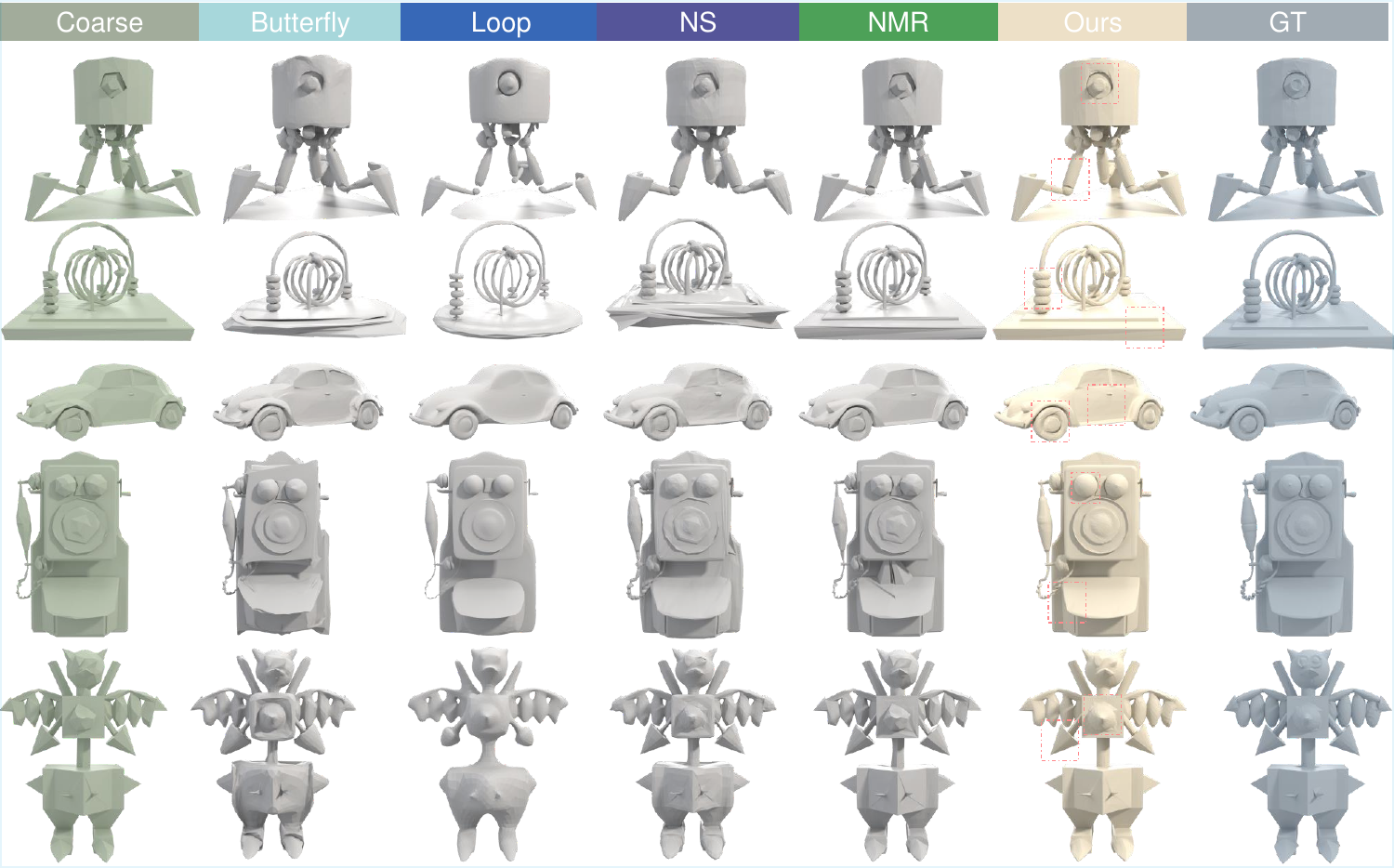} 
    \caption{\textbf{Qualitative comparison} We demonstrate the subdivision performance of different models on the same coarse meshes. Our model leverages global information to successfully recover fine details(highlighted in the boxes). While previous methods often misinterpret the subdivision structure, leading to over-smoothing or mesh artifacts, our approach effectively overcomes these issues. }
    \label{fig:sync_collapse}
    \vspace{-1.5em}
\end{figure}
\label{sec:experiment}
\subsection{Experimental Setup}

\vspace{-0.5em}
\noindent \textbf{Datasets.} 
We curate \textbf{FII-40k}, a high-quality manifold dataset refined from an initial 200k meshes sourced from 3D-FUTURE \cite{3dfuture}, G-Objaverse \cite{objaverse}, ShapeNet \cite{shapenet}, Toys4K \cite{stojanov2021using}, and TRELLIS-500K \cite{trellis}. The dataset is randomly split into training, validation, and testing sets following an 18:1:1 ratio.
\noindent \textbf{Baselines and Metrics.} 
We benchmark against traditional (Loop \cite{loop}, Modified Butterfly \cite{zorin1996interpolating}) and learning-based (Neural Subdivision \cite{NeurSu}, NMR \cite{NeurMeRe}) methods. Performance is evaluated via Chamfer Distance (CD), Hausdorff Distance (HD), and Normal Consistency \cite{fan2017point, cignoni1998metro}, computed on 300,000 points and averaged over two random seeds (40, 42) to reduce sampling variance.
\noindent \textbf{Implementation Details.}
Implemented in PyTorch \cite{paszke2019pytorch}, our model is trained on two RTX 4090 GPUs using AdamW \cite{loshchilov2017decoupled} ($LR=2\times10^{-4}$, batch size 128) for subdivision depth $L=2$. We set $T_{\text{init}}=2$ and $\lambda=0.5$ after activation. Random rotations are applied for augmentation.

\subsection{Results and Analysis}

\begin{table}[ht]
\centering
\caption{Quantitative comparison on the test set. Best results are highlighted in \textbf{bold}.}
\label{tab:comparison}
\small
\begin{tabular*}{\linewidth}{@{\extracolsep{\fill}} l ccc ccc @{}}
    \toprule
    \textbf{Data Type} & \multicolumn{3}{c}{\textbf{Closed Meshes Only}} & \multicolumn{3}{c}{\begin{tabular}[c]{@{}c@{}}\textbf{Full Dataset} \\ \textbf{(incl. Boundaries)}\end{tabular}} \\
    \cmidrule(lr){2-4} \cmidrule(lr){5-7}
    \textbf{Method} & \textbf{HD} $\downarrow$ & \textbf{CD} $\downarrow$ & \textbf{NC} $\uparrow$ & \textbf{HD} $\downarrow$ & \textbf{CD} $\downarrow$ & \textbf{NC} $\uparrow$ \\ 
    \midrule
    Butterfly & 0.07190 & 0.01138 & 0.9565 & 0.07880 & 0.01214 & 0.9492 \\
    Midpoint & 0.02706 & 0.01016 & 0.9718 & 0.03046 & 0.009818 & 0.9660 \\
    Loop & 0.06750 & 0.02003 & 0.9622 & 0.07208 & 0.01985 & 0.9558 \\
    NS & 0.03878 & 0.01337 & 0.9615 & 0.1067 & 0.01799 & 0.9434 \\
    NMR & 0.02386 & 0.00733 & 0.9799 & 0.06103 & 0.009357 & 0.9676 \\
    Ours & \textbf{0.01936} & \textbf{0.00629} & \textbf{0.9846} & \textbf{0.02328} & \textbf{0.006505} & \textbf{0.9793} \\ 
    \bottomrule
\end{tabular*}
\end{table}

\paragraph{Quantitative and Qualitative Performance.} 
As shown in Table \ref{tab:comparison}, our method consistently outperforms all baselines. On closed meshes, we reduce the HD of previous state-of-the-art (NMR) by 18.8\%. On the full dataset (including open surfaces), our method achieves 62\% lower HD (0.0233 vs. 0.0610) and 31\% lower CD (0.00651 vs. 0.00936) compared to adapted neural baselines. Qualitatively (Fig. \ref{fig:sync_collapse}), our model effectively captures global structure and fine details, whereas local neighborhood-based methods often suffer from structural collapse.

\paragraph{Handling Open Surfaces.} 
Unlike existing neural subdivision methods \cite{NeurSu, NeurMeRe} that are restricted to closed manifolds, our hybrid attention and boundary-aware features enable robust generalization to arbitrary topologies, providing a unified solution for complex open geometries.

\subsection{Ablation Study}

\begin{figure*}[t]
    \centering
    \newcommand{\figheight}{2.9cm} 
    
    \begin{subfigure}[b]{0.48\textwidth}
        \centering
        \includegraphics[height=\figheight]{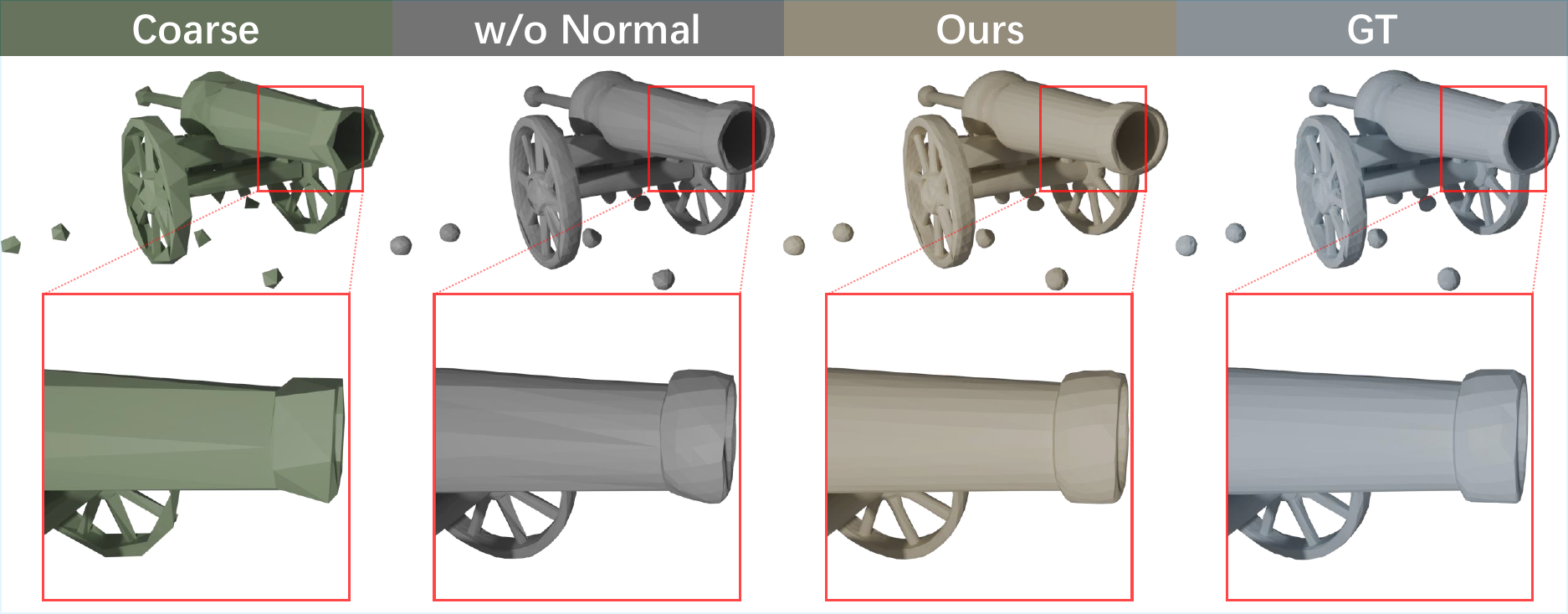}
        \caption{\textbf{Ablation on Normal}}
        \label{fig:normal}
    \end{subfigure}
    \hfill 
    \begin{subfigure}[b]{0.48\textwidth}
        \centering
        \includegraphics[height=\figheight]{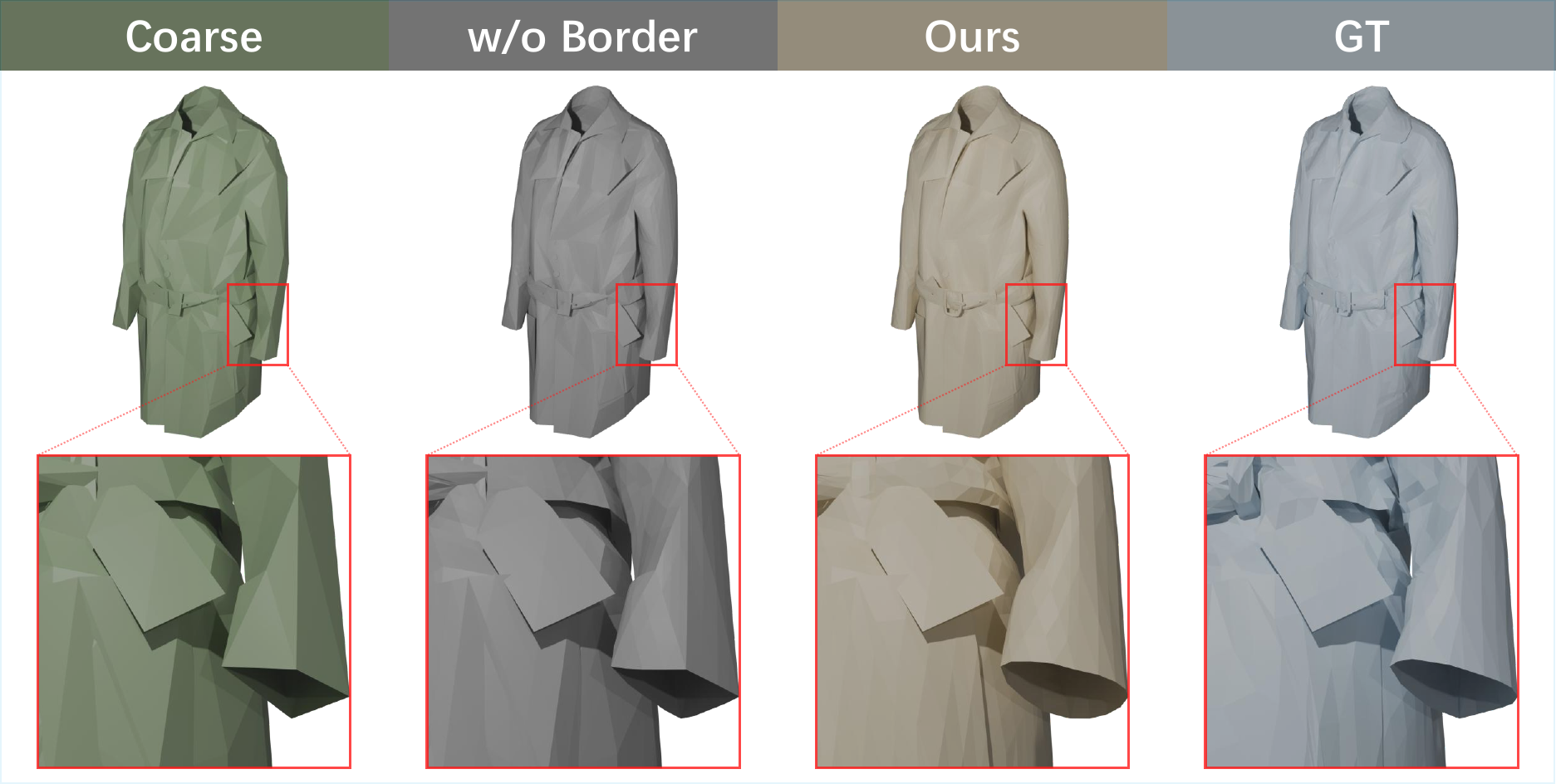}
        \caption{\textbf{Ablation on Border}}
        \label{fig:border}
    \end{subfigure}
    
    \caption{Ablation studies on different components. (a) Normal loss: The additional normal loss assists the model in accurately predicting vertex offsets, achieving superior geometric fidelity. (b) Border features: The boundary flag is crucial for our model, as the boundary status cannot be reliably inferred solely from vertex features or topological connectivity.}
    \label{fig:ablation_summary}
    \vspace{-2em}
\end{figure*}
\paragraph{Hybrid Attention Architecture.} 
Table~\ref{tab:abl_arch}a ablates the ratio of global to local attention layers using the "all" mode on the $40k \times 8$ rotated dataset. Our $2\times \text{global} + 6\times \text{local}$ configuration achieves optimal results, confirming that modest global context effectively guides local refinement. For the FII-40k scale, this hybrid design acts as a strong inductive bias that balances structural guidance with computational efficiency.

\paragraph{Supervision and Feature Components.} 
Table~\ref{tab:abl_comp}b evaluates the impact of training objectives and input features. Normal Loss ($\mathcal{L}_{norm}$) is essential for surface consistency and smoothness (\cref{fig:normal}), while the Border feature (boundary indicator) enables accurate representation of non-closed topologies (\cref{fig:border}). Their combination yields the highest reconstruction fidelity.
\begin{wrapfigure}[14]{r}{0.55\textwidth}
\centering
\includegraphics[width=\linewidth]{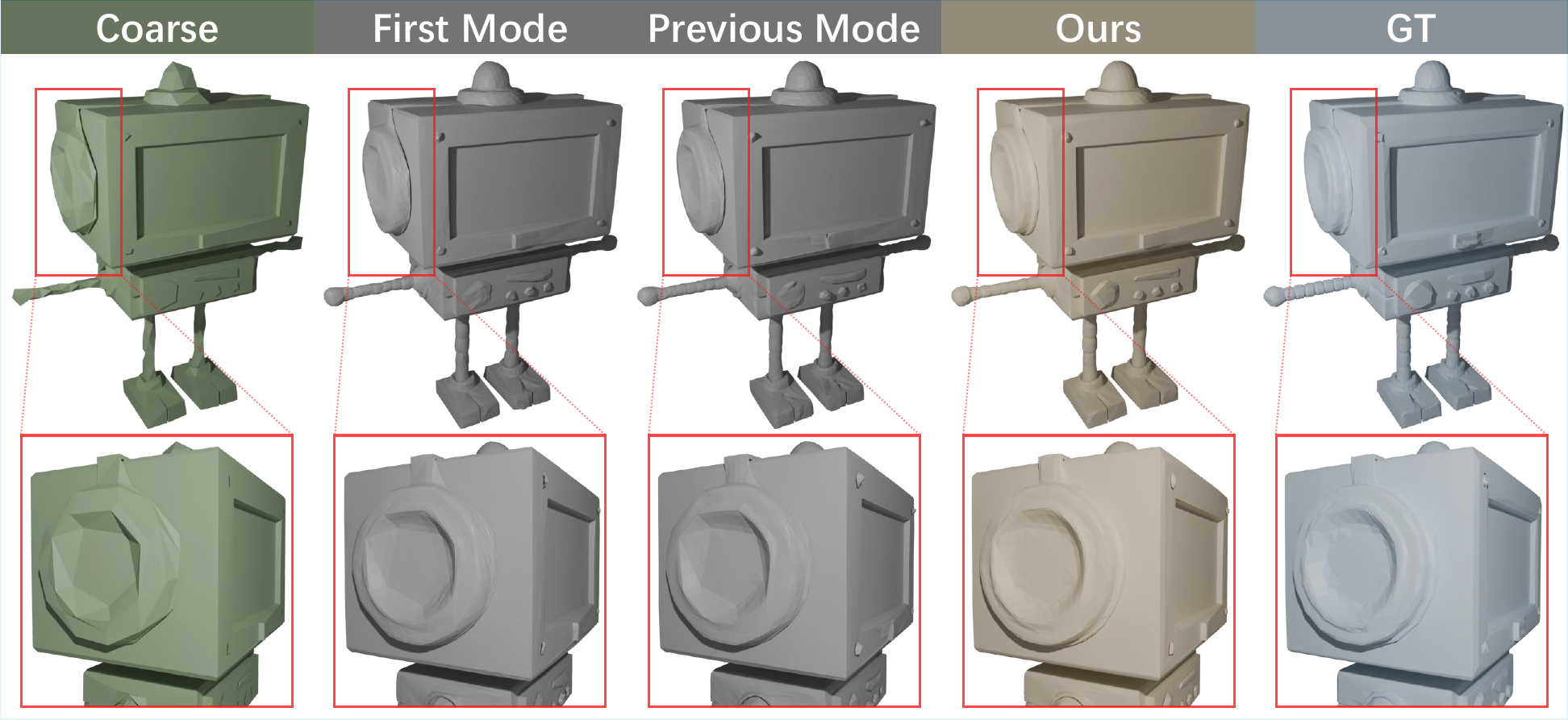}
\caption{\textbf{Ablation on Mode.} All mode achieves superior reconstruction quality, effectively recovering the complex structure of the robot's lateral auricle compared to other configurations on 80k scale.}
\label{fig:mask}
\end{wrapfigure}
\paragraph{Hierarchical Modes and Data Scaling.}
Table~\ref{tab:abl_mode}c investigates the synergy between the hierarchical receptive field and training data volume. The first\_cross and previous\_cross modes utilize the sampling-point self-attention query mechanism detailed in \cref{nscp}, while first and previous denote the hierarchical levels of the sampling points. (1) Mode Evolution: At a smaller scale (20k), the "first" mode acts as a strong regularizer that excels in stability. However, as the data scales up to 80k (
40k×2), the "all" mode becomes the superior strategy (see \cref{fig:mask}) by leveraging richer hierarchical context. (2) Scalability: Across all modes, doubling the training data significantly reduces geometric error, demonstrating the strong learning capacity and scalability of our model.
\begin{table}[htbp]
    \vspace{-1em}
    \centering
    \caption{Quantitative ablation studies on different components.}
    \label{tab:ablation_left_right}
    
    \begin{minipage}[t]{0.42\textwidth}
        \centering
        (a) Architecture analysis 
        \vspace{0.1cm}
        \resizebox{\linewidth}{!}{
        \begin{tabular}{lccc}
        \toprule
        \textbf{Arch} & \textbf{HD} & \textbf{CD} & \textbf{NC} \\ \midrule
        Global & 0.02373 & 0.006751 & 0.9778 \\
        Local & 0.02365 & 0.006642 & 0.9786 \\
        \textbf{Hybrid} & \textbf{0.02328} & \textbf{0.006505} & \textbf{0.9793} \\ \bottomrule
        \end{tabular}
        \label{tab:abl_arch}
        }
        
        \vspace{0.1cm} 
        
        (b) Supervision and Features \label{tab:abl_comp} \\ \vspace{0.1cm}
        \resizebox{\linewidth}{!}{
        \begin{tabular}{lccc}
        \toprule
        \textbf{Setting} & \textbf{HD} & \textbf{CD} & \textbf{NC} \\ \midrule
        None & 0.02420 & 0.006722 & 0.9769 \\
        Normal Loss & 0.02403 & 0.006566 & 0.9781 \\
        Border Feat. & 0.02417 & 0.006633 & 0.9773 \\
        \textbf{Both} & \textbf{0.02361} & \textbf{0.006560} & \textbf{0.9783} \\ \bottomrule
        \end{tabular}
        }
    \end{minipage}\hfill 
    \begin{minipage}[t]{0.54\textwidth}
        \centering
        (c) Impact of hierarchical modes and training data scale\label{tab:abl_mode} \\ \vspace{0.1cm}
        \resizebox{\linewidth}{!}{
        \begin{tabular}{lcccc}
        \toprule
        \textbf{Mode} & \textbf{Data} & \textbf{HD} & \textbf{CD} & \textbf{NC} \\ \midrule
        first\_cross & 20k & 0.02897 & 0.007832 & 0.9640 \\
        prev\_cross & 20k & 0.02848 & 0.007346 & 0.9706 \\
        first & 20k & 0.02540 & 0.006991 & 0.9738 \\
        prev & 20k & 0.02582 & 0.007020 & 0.9726 \\
        all & 20k & 0.02611 & 0.007063 & 0.9760 \\ \midrule
        first & 40k*2 & 0.02486 & 0.006772 & 0.9768 \\
        prev & 40k*2 & 0.02420 & 0.006722 & 0.9769 \\
        \textbf{all} & \textbf{40k*2} & \textbf{0.02395} & \textbf{0.006640} & \textbf{0.9783} \\ \bottomrule
        \end{tabular}
        }
    \end{minipage}
    \vspace{-1em}
\end{table}

\section{Discussion and Limitations}

Despite its robustness, our method has two primary limitations. 
First, a domain gap exists between synthetic training pairs and real-world modeling scenarios. While our pipeline mitigates noise, automated decimation still cannot fully replicate the nuanced edge-flow of artist-authored cages. This discrepancy, along with residual discretization artifacts, creates a ``learning ceiling'' for extremely fine details, which could be further addressed by utilizing more advanced, topology-preserving simplification algorithms to generate training data.
Second, using deterministic $L_2$ regression to solve a one-to-many inverse problem leads to over-smoothing. Because the model predicts the statistical mean of possible high-resolution geometries, it tends to produce ``blurred'' results and fails to recover sharp components that disappeared during decimation, despite the 1-to-4 subdivision structure providing sufficient degrees of freedom.Incorporating generative modeling objectives represents a promising direction to alleviate this effect.

\section{Conclusion}

We presented a robust neural subdivision framework SubdivAR combining a curated large-scale dataset with a Hybrid Topology-Aware Transformer. Our approach significantly outperforms both traditional rules and previous learning-based methods, particularly in handling complex open surfaces. Our analysis suggests that the primary bottleneck remains the information loss during simplification and the smoothing nature of regression losses. Future research will explore generative modeling to overcome these limitations and fully reconstruct non-trivial geometric features lost in coarse representations.

\par\vfill\par

\bibliography{main}
\bibliographystyle{plainnat}
\appendix

\section{Dataset}

\subsection{Data Collection}

We collect meshes from several public datasets and apply a unified preprocessing and filtering pipeline to ensure geometric fidelity, structural integrity, and shape diversity. Table~\ref{tab:data_stats} summarizes the number of valid objects retained after preprocessing.

\begin{wraptable}{r}{0.38\linewidth}
\vspace{-28pt}
\centering
\begin{tabular}{lc}
\toprule
Dataset & \#Objects \\
\midrule
3D-FUTURE & 924 \\
g-objaverse & 22,332 \\
ShapeNetCore.v2 & 1,425 \\
Toys4K & 673 \\
TRELLIS-500 subset& 19,535 \\
\midrule
Total & 44,889 \\
\bottomrule
\end{tabular}
\caption{\textbf{Dataset statistics after preprocessing.} We report the number of valid objects retained from each source repository(covers diverse models like furniture and animals). Filtering pass rates varied significantly across datasets ; notably, only a subset of Objaverse was utilized.}
\label{tab:data_stats}
\end{wraptable}

\begin{algorithm}[b]
\caption{Mesh Normal Orientation Correction}
\label{alg:normal_fix}
\begin{algorithmic}[1]
\Require vertices $V$, faces $F$
\Ensure corrected mesh $(V',F')$
\State $(F',C) \gets \text{BFSOrient}(F)$
\State $V' \gets \text{Normalize}(V)$
\State $M \gets (V',F')$
\State Sample cameras $\{c_i\}_{i=1}^{N_c}$ on sphere
\For{each $c_i$}
    \State Sample points $\{p_j\}$ in bounding box
    \State $r_j \gets p_j - c_i$
    \State $f_j \gets \text{RayFirst}(c_i,r_j)$
    \If{$f_j \neq -1$}
        \State $d \gets \langle n_{f_j}, r_j \rangle$
        \State vote$(f_j) \mathrel{+}= \text{sign}(d)$
    \EndIf
\EndFor
\For{component $k$}
    \If{$\sum_{f\in k}$ vote$(f) < 0$}
        \State flip faces in $k$
    \EndIf
\EndFor
\State \Return $(V',F')$
\end{algorithmic}
\end{algorithm}
\subsection{Preprocessing} 
\label{sec:datapreocess}

The raw meshes are processed through a standardized pipeline:
\begin{enumerate} 
    \item \textbf{Decimation and Cleaning:} Filter meshes with $<2,500$ faces. Models $>40,000$ faces are simplified via QEM (\citet{qem}). We perform cleaning and repair non-manifold structures using standard mesh processing libraries (\citet{zhou2018pymesh}).
    \item \textbf{Orientation and Normalization:} Face orientations are corrected using view‑dependent visibility methods(\cref{alg:normal_fix}) and meshes are normalized into a $[-1, 1]$ canonical space.
    \item \textbf{Subdivision Generation:} Cleaned meshes undergo  stochastic decimation (\citet{NeurSu}) to a coarse resolution of $750 \text{--} 850$ faces and re-parametrization.
\end{enumerate} 
\subsection{Filtering}
\begin{figure}[t]
    \centering
    \includegraphics[width=1.0\linewidth]{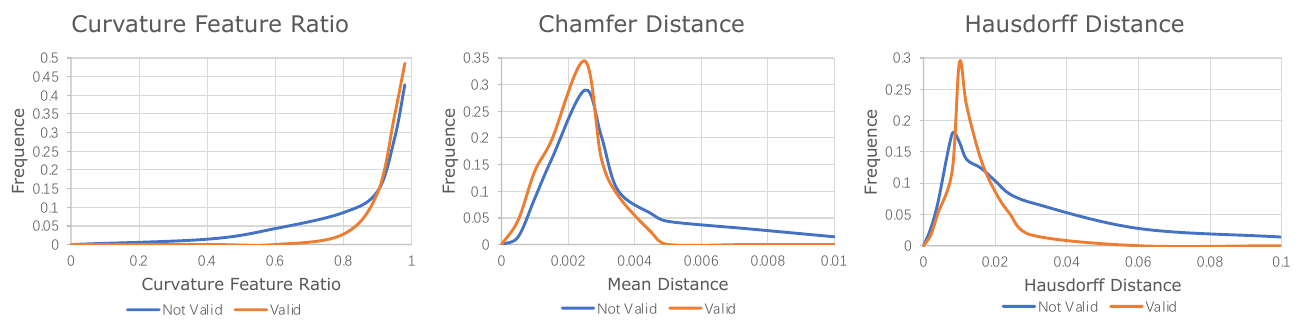} 
    \caption{\textbf{Metric Distribution} We visualize the metric distributions for both successful and failed samples. Because failed samples are selected based on subpar performance in specific dimensions, their distributions inevitably overlap with those of the successful ones. Furthermore, the failed samples demonstrate a broader distribution with prominent long-tail effects.}
    \label{fig:metric}
\end{figure}

To further guarantee geometric fidelity and structural integrity, mesh pairs are evaluated using a set of geometric and topological metrics, as summarized completely in Table~\ref{tab:metrics_details2}. Thresholds are chosen to balance accuracy, topology integrity, and shape diversity. Distrubtion of some metric between successful and failed samples is shown in \cref{fig:metric}

\begin{table}[htbp]
\centering
\caption{\textbf{Detailed metrics for data curation.} Thresholds balance fidelity, topology integrity, and shape diversity.}
\label{tab:metrics_details2}
\begin{tabular}{@{}lccccc@{}}
\toprule
Category & Metric ($m_i$) & Scope & Objective & $T_i$ & Valid Condition \\ 
\midrule
Fidelity 
& HD ($sub \rightarrow ori$) & $M_{sub} \rightarrow M_{orig}$ & Accuracy & 0.028 & $m_i < T_i$ \\
& HD ($ori \rightarrow sub$) & $M_{orig} \rightarrow M_{sub}$ & Accuracy & 0.035 & $m_i < T_i$ \\
\midrule
Integrity 
& Vertex/Face Ratio & $M_{sub}$ & Structure & 0.55 & $m_i < T_i$ \\
& Self-intersection & $M_{orig}$ & Topology & 0.15 & $m_i < T_i$ \\
& Self-intersection & $M_{sub} \leftarrow M_{orig}$ & Topology & 0.08 & $|m_i| < T_i$ \\
\midrule
Informativeness 
& Edge Feature Ratio & $M_{orig}$ & Diversity & 0.025 & $m_i < T_i$ \\
\bottomrule
\end{tabular}
\end{table}

\section{Methodology}
\subsection{Network Architecture}

Our network first applies a 48-dimensional sinusoidal positional encoding, followed by eight attention modules with a hidden size of 512, including two global memory attention layers and six local attention layers, all operating in the \textit{all} mode. In our ablation studies, all hierarchical attention modes consistently utilize the same eight-layer attention module structure, with the only variable being the scope of their respective receptive fields. The ground-truth vertex offsets are scaled by a factor of 100 during training.

\subsection{Training Details}
All models are trained using the Adam optimizer with a learning rate of $2\times 10^{-4}$ and a total batch size of 128, utilizing gradient accumulation over 8 steps. Training typically runs for 20 epochs on two NVIDIA RTX 4090 GPUs.

\subsection{Comparison with Alternative Refinement Paradigms}
As discussed in the related work, various 3D super-resolution (SR) paradigms exist beyond the mesh-based subdivision used in this work, yet their functional goals and applicability differ significantly.

\paragraph{Point Cloud Upsampling} 
Point cloud upsampling (PCU) is often perceived as a potential alternative for surface enhancement, yet its behavior is highly sensitive to input sampling density. If a coarse mesh is over-sampled to provide the initial point set, the PCU model tends to strictly adhere to the original faceted geometry, merely increasing the density of points within flat planes rather than smoothing them into a curved manifold. This occurs because the points sampled from the faces act as interference, locking the model into the existing coarse structure. Conversely, if only sparse samples are used—ideally restricted to the mesh vertices to eliminate face-point interference—the task begins to align with subdivision-style geometric extrapolation. 

\paragraph{Implicit and Voxel-based Methods.} 
Methods leveraging implicit representations and voxel upsampling can indeed generate smoother surfaces by increasing volumetric resolution. While they can achieve results visually similar to subdivision, they suffer from a major workflow bottleneck: they produce unorganized, excessively dense meshes. Such outputs are incompatible with standard artistic and industrial modeling pipelines, as they lack the structured edge-flow required for downstream tasks.

\paragraph{Generative and Autoregressive Mesh Models.} 
Recent generative frameworks like MeshRipple \cite{lin2025meshripple} and LATO \cite{zhao2026lato} attempt to decode structured meshes from high-resolution voxel features. While promising, these methods currently face significant hurdles, including incomplete modeling, non-manifold holes, and the loss of fine-grained details, which hinder their practical adoption. 
Alternatively, the autoregressive approach, exemplified by AR-Mesh \cite{ARMesh}, treats the generation process as a reverse collapse of a coarse mesh from the origin. While this allows for simultaneous refinement of topology and vertex positions, its limitations are severe and difficult to resolve. Primarily, it is constrained by scalability, with output resolution typically capped at approximately 1,000 faces—far below the precision required for high-resolution refinement. Furthermore, due to the inherent sequential nature of autoregressive sampling, inference time grows linearly with sequence length, often requiring several minutes or longer to generate a single mesh.
\section{Additional Experiments}
\begin{figure}[htbp]
    \centering
    \includegraphics[width=\linewidth]{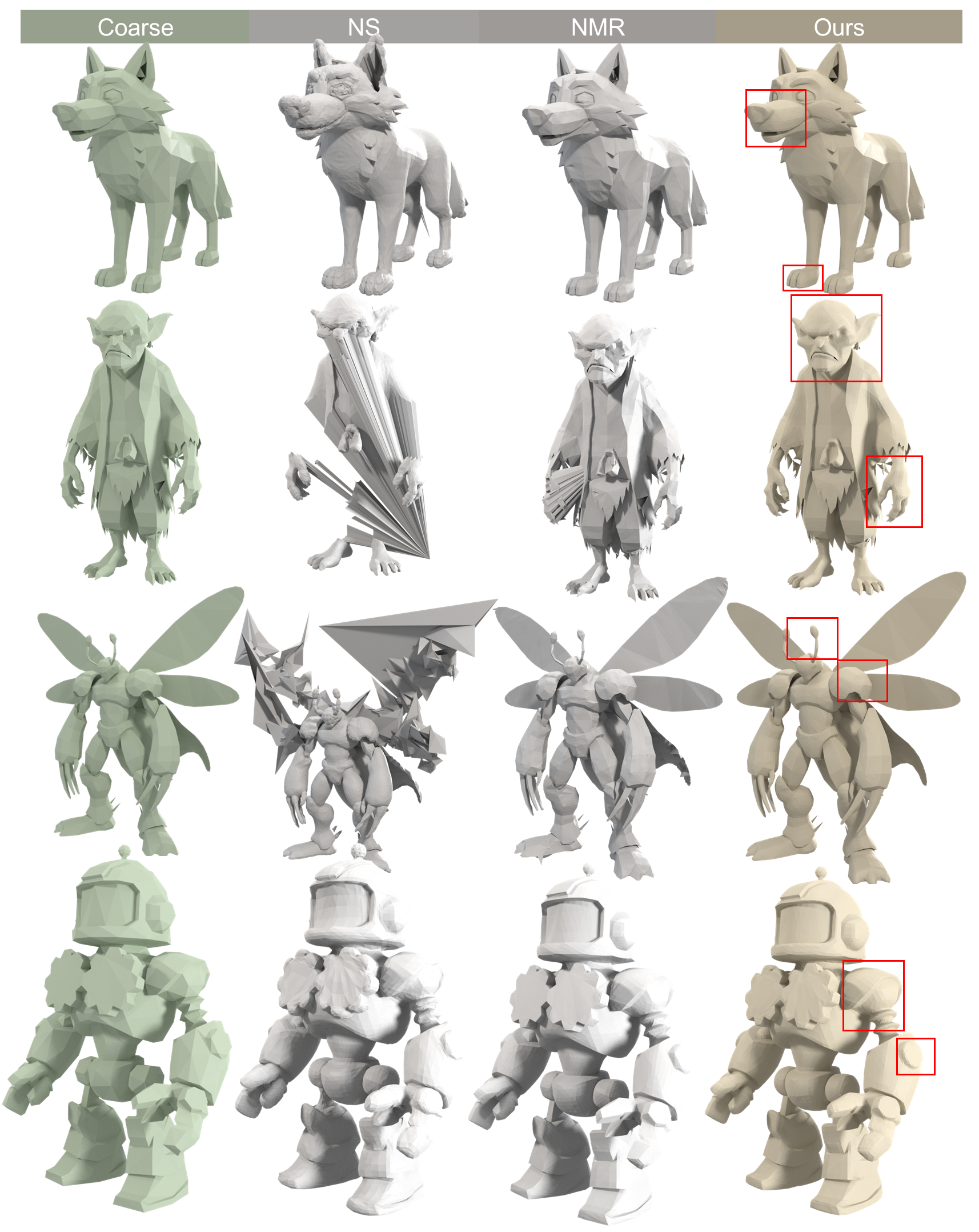}
    \caption{\textbf{Test on coarse mesh on generated by Tripop1.0} The primary failure mode for several baseline models, such as NS \cite{NeurSu} and NMR \cite{NeurMeRe}, stems from their inherent inability to process meshes with open boundaries. To facilitate a visual comparison on such geometries, we manually pre-processed these meshes by stitching all boundary vertices to a singular auxiliary point. Furthermore, our model yields encouraging preliminary results on meshes with regularized topologies from generative models (e.g., Tripo 1.0), despite such structures being absent from our training set. This suggests a strong potential for generalization; with further training on datasets that include diverse regularized topologies and varying resolutions, SubdivAR could potentially serve as a flexible refiner within emerging AI-driven modeling workflows.}
    \label{fig:on_generative}
\end{figure}
As Neural Progress Meshes and Grape Neural Subdivision are not publicly available, we include additional ablation experiments for further analysis.\cref{tab:attention_arch} compares the performance of different attention architectures under varying training data scales on "all" mode.\cref{tab:local_ring} evaluates the effect of different local attention neighborhoods under the all mode.
\cref{fig:on_generative} compares the subdivision results on coarse meshes that were not generated via standard QEM decimation \cite{qem}. Instead, these inputs are sourced from Tripop1.0 , which produces meshes with more regularized topologies and face counts ranging from 1,000 to 2,000. Despite the significant discrepancy in both topology and resolution compared to our training distribution, SubdivAR consistently yields high-quality outputs, further demonstrating its cross-domain robustness. These results suggest that by utilizing topology-regular simplification algorithms for data synthesis and expanding the dataset to encompass a wider range of input scales, our framework could be seamlessly integrated into professional mesh design pipelines, significantly alleviating the manual labor required for high-fidelity refinement.

In terms of inference speed, as shown in \cref{tab:time_horizontal}, NS \cite{NeurSu} and NMR \cite{NeurMeRe} exhibit the lowest latencies (0.0093s and 0.0126s, respectively) due to their lightweight local operators. SubdivAR entails a higher computational cost of 0.0547s per mesh, primarily attributed to the Transformer-based global semantic modeling. Nevertheless, our inference latency remains well under 0.1s, which is sufficient for real-time feedback in interactive modeling workflows. It is anticipated that even when scaling to larger model capacities or training on more extensive, topologically consistent datasets, the execution time of SubdivAR will remain highly acceptable, ensuring its feasibility in professional production environments.
\begin{table}[htbp]
\centering
\caption{\textbf{Comparison of attention architectures under different training data scales.}}
\label{tab:attention_arch}
\begin{tabular}{lcccccc}
\toprule
& \multicolumn{3}{c}{20k} & \multicolumn{3}{c}{40k $\times$ 8} \\
\cmidrule(lr){2-4} \cmidrule(lr){5-7}
Arch & HD $\downarrow$ & CD $\downarrow$ & NC $\uparrow$ 
     & HD $\downarrow$ & CD $\downarrow$ & NC $\uparrow$ \\
\midrule
Global & 0.03217 & 0.008532 & 0.9603
       & 0.02373 & 0.006751 & 0.9778 \\

Local  & \textbf{0.02542} & \textbf{0.007062} & \textbf{0.9760}
       & 0.02365 & 0.006642 & 0.9786 \\

Hybrid & 0.03085 & 0.008294 & 0.9617
       & \textbf{0.02328} & \textbf{0.006505} & \textbf{0.9793} \\
\bottomrule
\end{tabular}
\end{table}
\begin{table}[h]
\centering
\caption{\textbf{Ablation on local attention neighborhoods in the all mode.}}
\label{tab:local_ring}
\begin{tabular}{lccc}
\toprule
Neighborhood & HD $\downarrow$ & CD $\downarrow$ & NC $\uparrow$ \\
\midrule
1-ring & \textbf{0.02328} & \textbf{0.006505} & \textbf{0.9793} \\
2-ring & 0.02357 & 0.006607 & 0.9788 \\
\bottomrule
\end{tabular}
\end{table}
\begin{table}[h]
\centering
\caption{Inference latency comparison (seconds per mesh).}
\label{tab:time_horizontal}
\begin{tabular}{lccc}
\toprule
Method & NS~\cite{NeurSu} & NMR~\cite{NeurMeRe} & SubdivAR (Ours) \\
\midrule
Avg. Time (s) $\downarrow$ & 0.0093 & 0.0126 & 0.0747 \\
\bottomrule
\end{tabular}
\end{table}
\section{More Qualitative Results}
\begin{figure}[htbp]
    \centering
    \resizebox{1\linewidth}{!}{
    \includegraphics[width=1\linewidth]{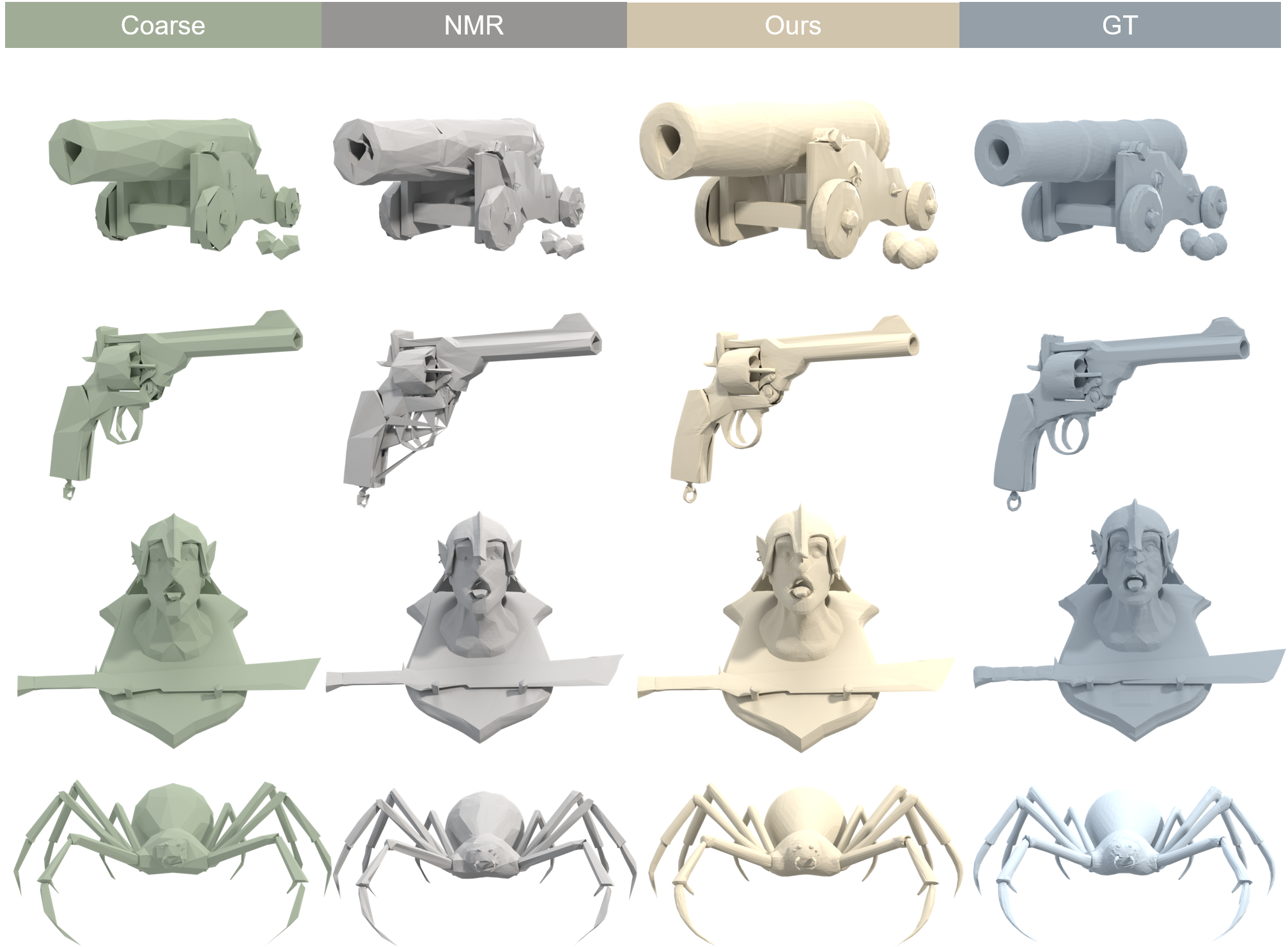} 
    }
    \caption{\textbf{More Qualitative comparison} We demonstrate the subdivision performance of different models on the same coarse meshes.}
    \label{fig:more Qualitative comparison}
\end{figure}

\begin{figure}[htbp]
    \centering

    \includegraphics[width=1\linewidth]{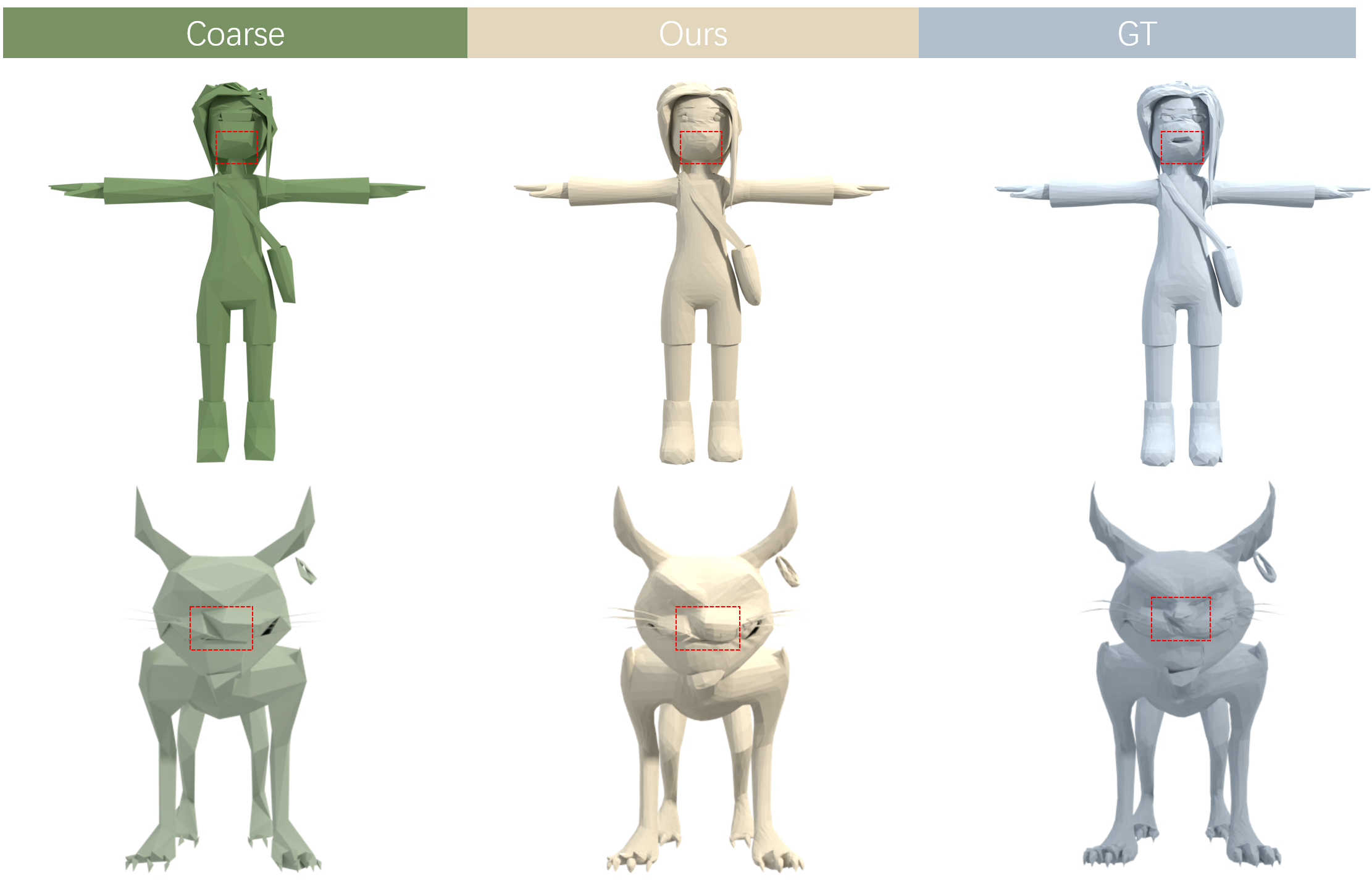} 

    \caption{\textbf{Failure cases} Our method fails to recover sufficient geometric details in these examples. The missing details are highlighted with red boxes.}
    \label{fig:failure case}
\end{figure}

Additional qualitative comparisons are provided in \cref{fig:more Qualitative comparison}, where SubdivAR demonstrates a superior ability to leverage global semantic context to guide the precise placement of local vertices. In \cref{fig:failure case}, we analyze several failure cases where the model failed to fully recover the complex structures of the original meshes. These limitations primarily stem from the inherent constraints of deterministic mean-value prediction and the current scale of our dataset; we acknowledge that 40,000 training pairs are likely insufficient for modern high-capacity models to learn an exhaustive range of geometric priors. 

\begin{figure}[htbp]
    \centering
    \includegraphics[width=\linewidth]{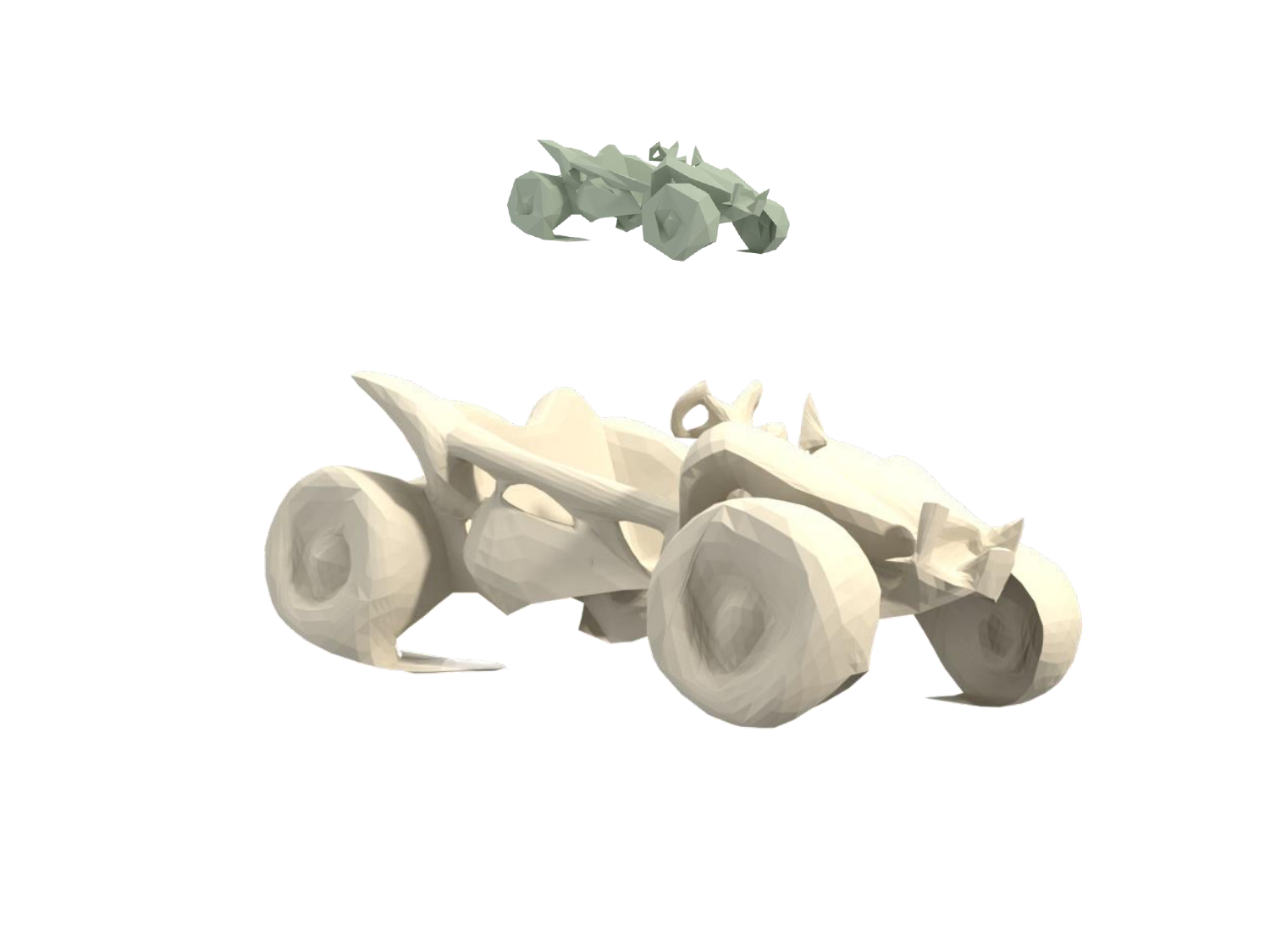}
    \caption{\textbf{More results}}
\end{figure}

\begin{figure}[htbp]
    \centering
    \includegraphics[width=0.9\linewidth]{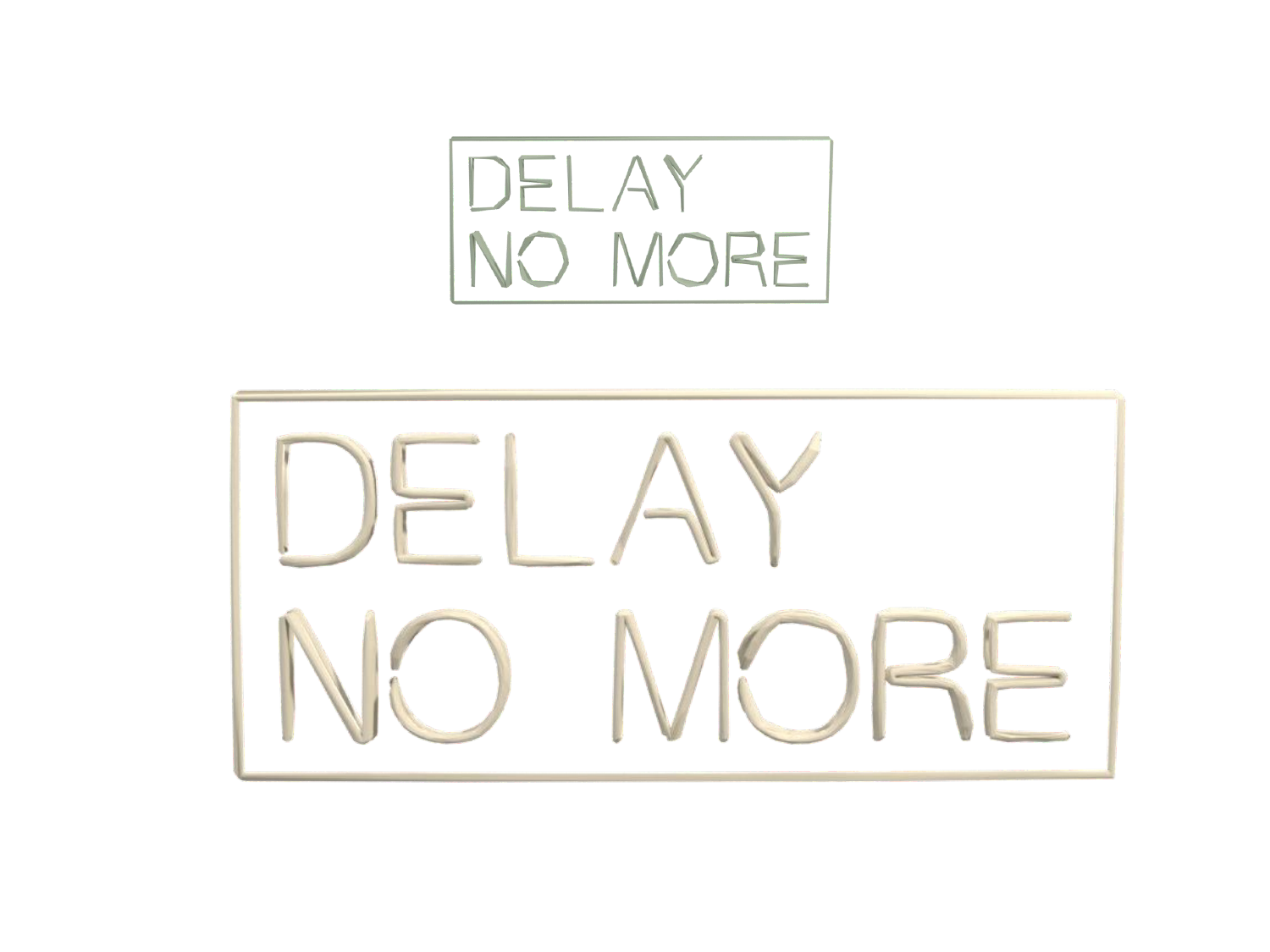}
    \caption{\textbf{More results}}
\end{figure}

\begin{figure}[htbp]
    \centering
    \includegraphics[width=0.9\linewidth]{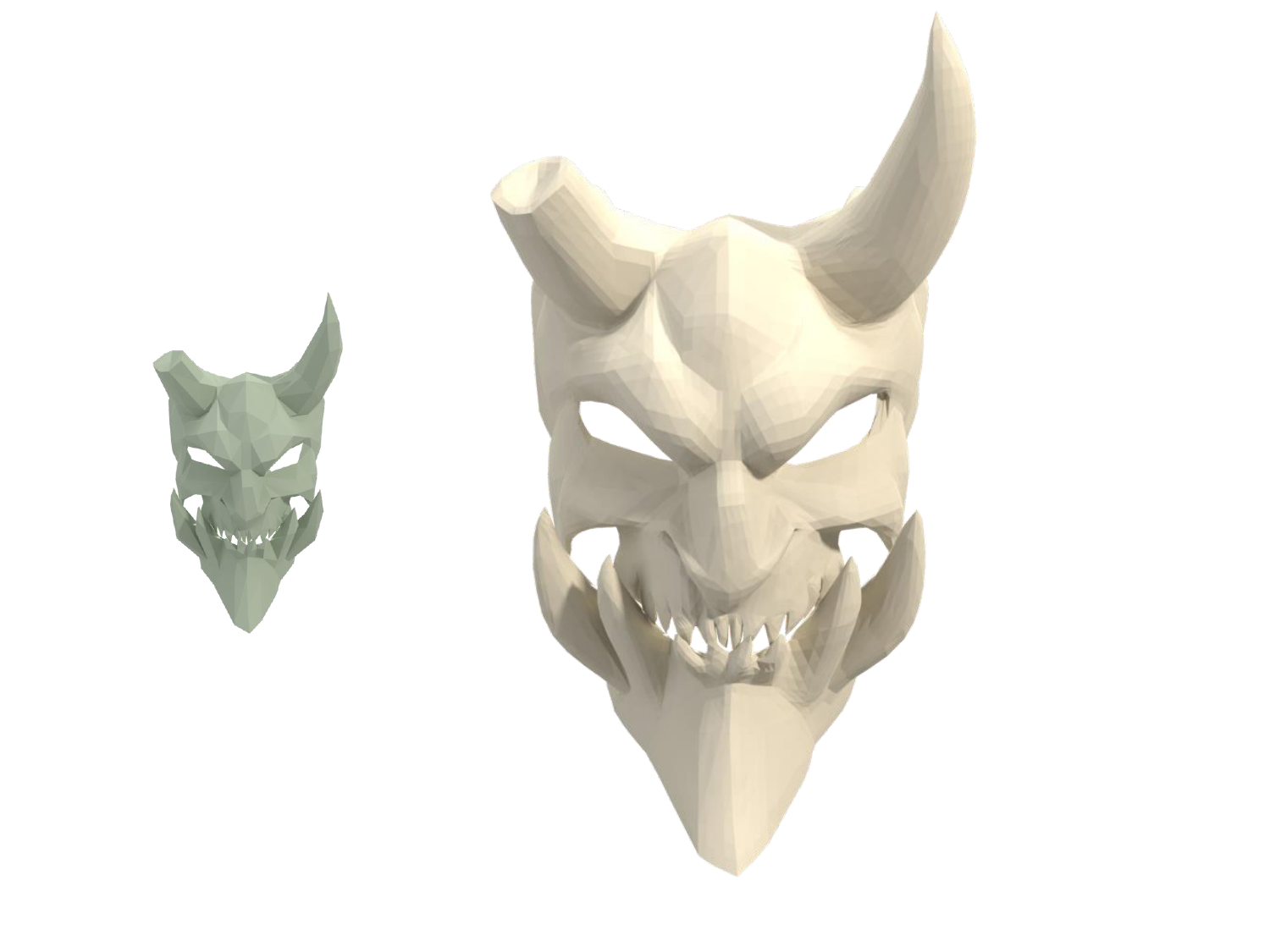}
    \caption{\textbf{More results}}
\end{figure}

\begin{figure}[htbp]
    \centering
    \includegraphics[width=0.9\linewidth]{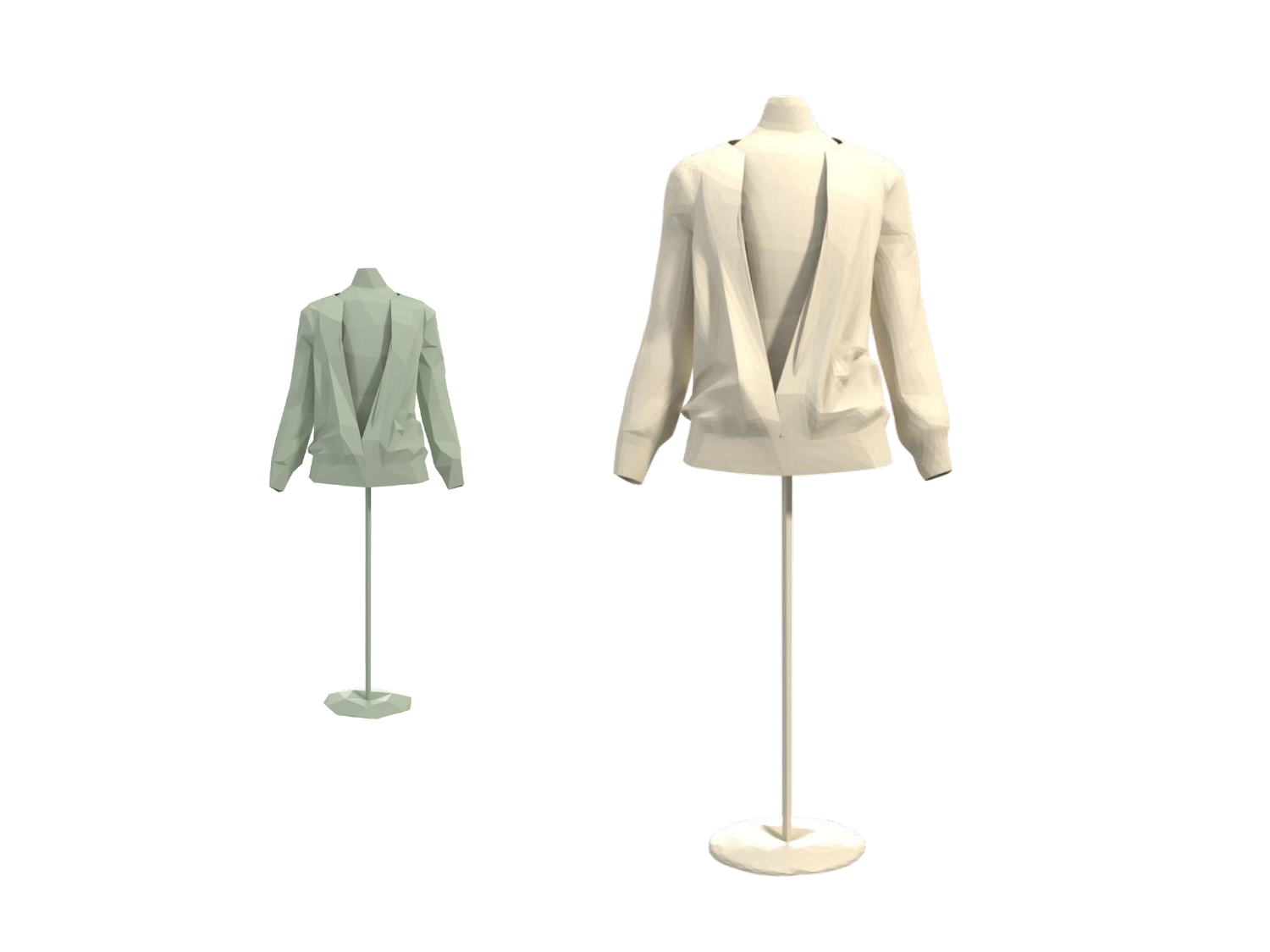}
    \caption{\textbf{More results}}
\end{figure}

\begin{figure}[htbp]
    \centering
    \includegraphics[width=\linewidth]{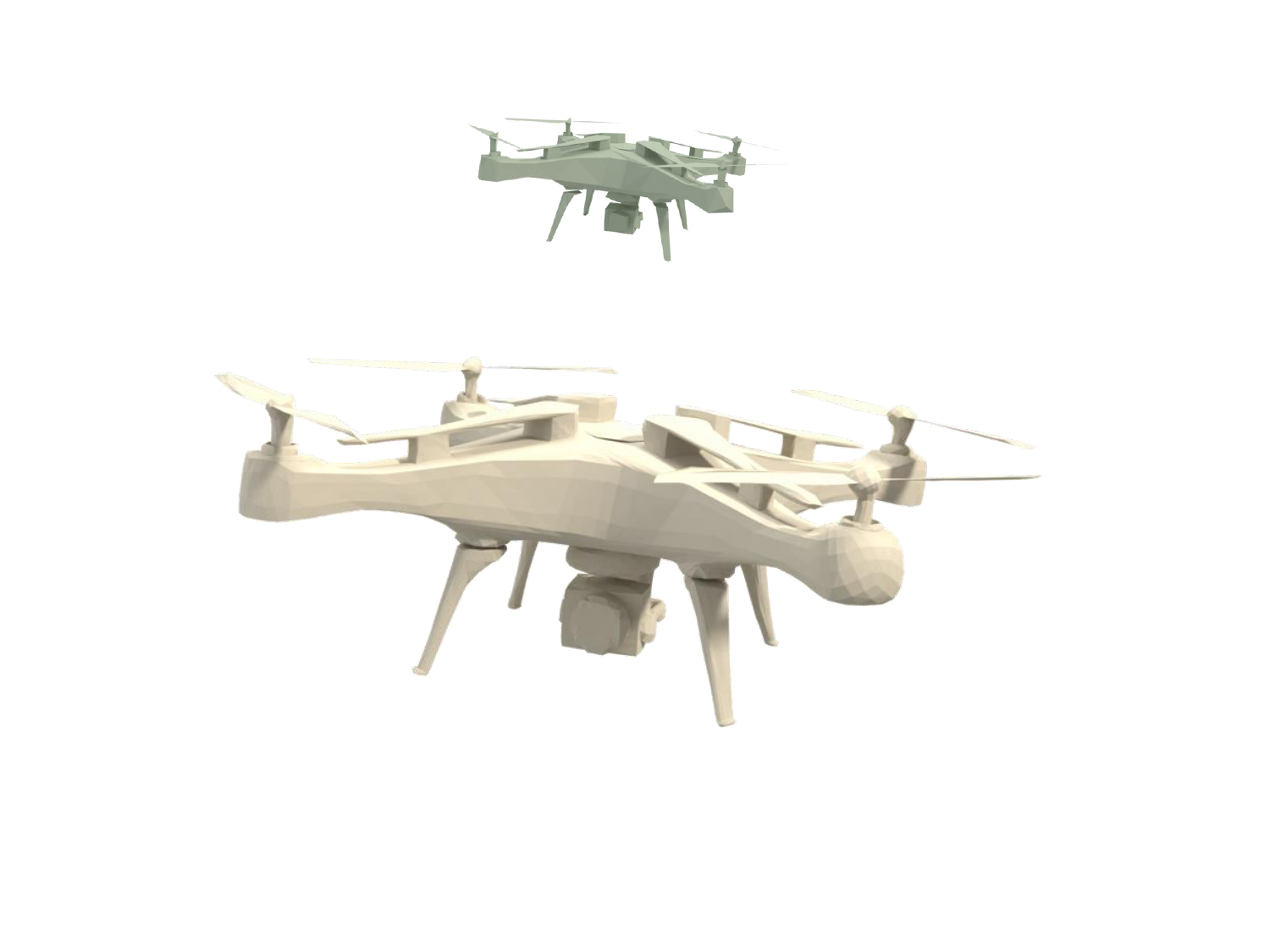}
    \caption{\textbf{More results}}
\end{figure}

\begin{figure}[htbp]
    \centering
    \includegraphics[width=\linewidth]{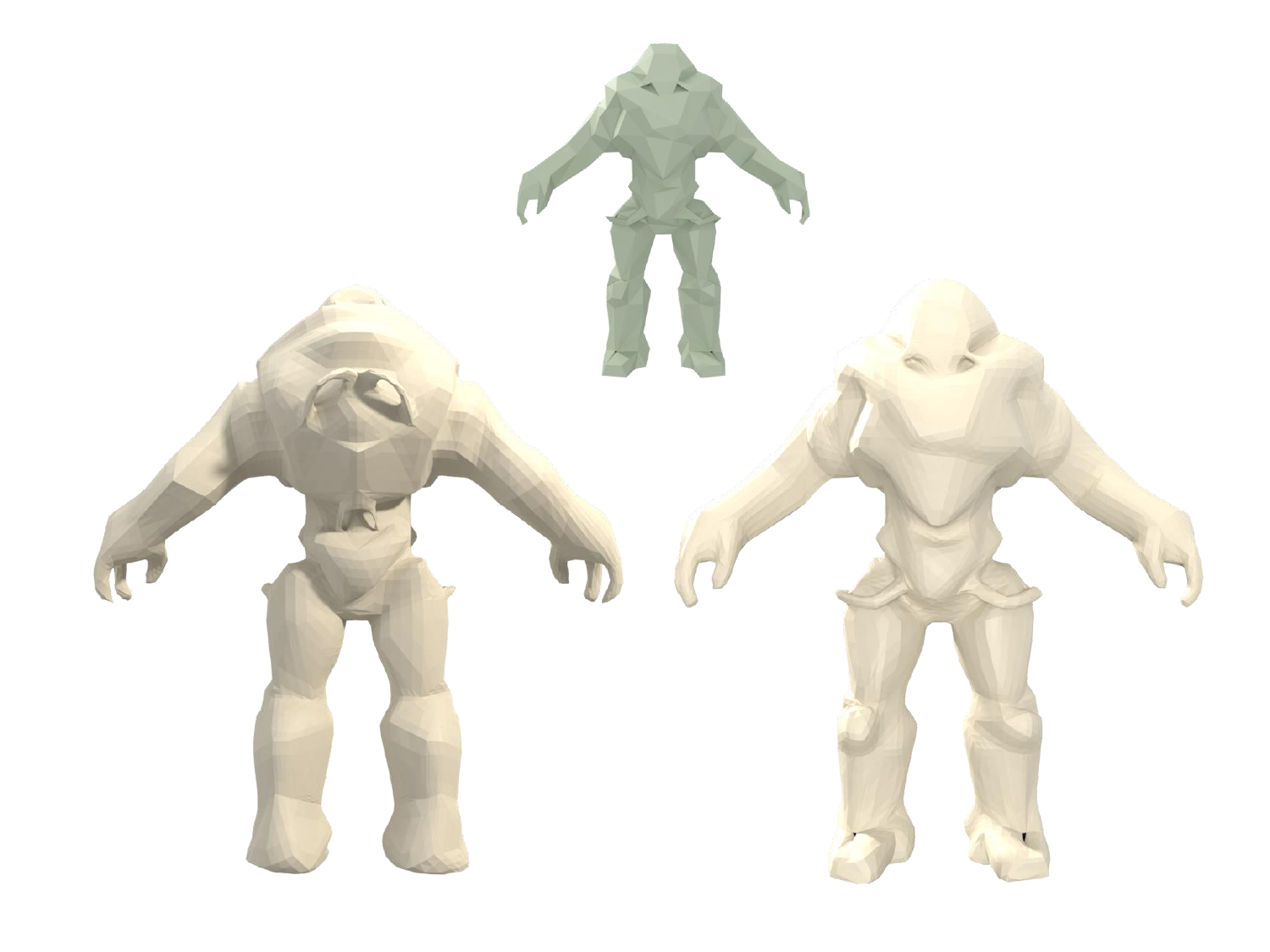}
    \caption{\textbf{More results}}
\end{figure}

\end{document}